\documentclass[10pt,journal,compsoc]{IEEEtran}
\usepackage{graphicx}%
\usepackage{multirow}%
\usepackage{float}
\usepackage{lineno}
\usepackage{amsmath,amssymb,amsfonts}%
\usepackage{amsthm}%
\usepackage{mathrsfs}%
\usepackage{ragged2e}
\usepackage{xcolor}%
\usepackage{booktabs}
\usepackage{makecell}
\usepackage{pifont}
\usepackage{graphicx}
\usepackage{textcomp}%
\usepackage{manyfoot}%
\usepackage{booktabs}%
\usepackage{algorithm}%
\usepackage{algorithmicx}%
\usepackage{algpseudocode}%
\usepackage{listings}%
\usepackage{multirow}
\usepackage{siunitx}
\usepackage{subcaption}
\usepackage{placeins}
\usepackage[english]{babel}
\usepackage{float}

\usepackage{array}
\usepackage{arydshln}
\usepackage{xspace}
\newcommand{\ie}{{\emph{i.e.}},\xspace}

\theoremstyle{plain}

\theoremstyle{remark}

\theoremstyle{definition}

\begin{document}

\title{Beyond Forgetting in Continual Medical Image Segmentation: A Comprehensive Benchmark Study} 



%



\author{Bomin~Wang$^{\dagger}$,
        Hangqi~Zhou$^{\dagger}$,
        Yibo~Gao,
        and~Xiahai~Zhuang$^*$%
\IEEEcompsocitemizethanks{
\IEEEcompsocthanksitem B. Wang, H. Zhou, Y. Gao and X. Zhuang are with the School of Data Science, Fudan University, Shanghai, China.
\IEEEcompsocthanksitem $^\dagger$ B. Wang and H. Zhou contributed equally to this work.
\IEEEcompsocthanksitem $^*$ X. Zhuang is the corresponding author.}%
}

\markboth{Journal of \LaTeX\ Class Files,~Vol.~14, No.~8, August~2015}%
{Shell \MakeLowercase{\textit{et al.}}: Bare Demo of IEEEtran.cls for Computer Society Journals}

\IEEEtitleabstractindextext{%
\begin{abstract}

Continual learning (CL) is essential for deploying medical image segmentation models in clinical environments where imaging domains, anatomical targets, and diagnostic tasks evolve over time. 
However, continual segmentation still faces three main challenges. First, the scenarios for this task remain insufficiently standardized for real-world clinical settings. Second, existing research has been primarily focused on mitigating forgetting, overlooking the other essential properties such as plasticity. Third, a benchmark work with comprehensive evaluation on existing methods is stll desirable. 
To address these gaps, we present such benchmark study of continual medical image segmentation. 
We first define three clinically motivated scenarios, namely Domain-CL, Class-CL, and Organ-CL, to respectively capture the cross-center domain shift, the incremental anatomical structure segmentation, and the cross-organ segmentation.
We then introduce an evaluation framework that measures not only general performance and forgetting, but also plasticity, forward generalizability, parameter efficiency, and replay burden.
The results, from extensive experiments with representative CL methods, showed that it was still challenging to develop a model that could satisfy all the requirements simultaneously. 
Nevertheless, these studies also suggested that the replay-based methods achieve the best overall balance between stability and plasticity, 
the parameter-isolation methods should be  effective at reducing forgetting, though at the cost of increased model size, 
and the forward generalizability remain a significantly understudied aspect of this research field. 
Finally, we discuss related learning paradigms and outline future directions for continual medical image segmentation.
\end{abstract}

}

\maketitle
\IEEEdisplaynontitleabstractindextext

\section{Introduction}\label{sec1}

The past decade has witnessed the rapid rise of artificial intelligence (AI) in medicine \cite{back_AI_for_medicine, back_AI_for_medicine_retina}.
However, most deployed medical AI systems cannot acquire new knowledge when faced with new data or tasks. By contrast, human experts continue to learn throughout their professional practice \cite{back_distribution_shift_nips, back_distribution_shift_biostat}. 
For instance, a well-trained pulmonologist could continually identify emerging COVID-19 variants (e.g., Alpha, Delta, Omicron). In stark contrast, AI models cannot effectively learn new features to diagnose new variants \cite{back_AI_for_COVID}.  
This limitation arises because most current AI models are trained in a static manner and their parameters remain fixed after training.
Recently, continual learning (CL) has emerged as a paradigm that allows AI systems to acquire new knowledge incrementally.
Therefore, CL is key to realizing the full potential of medical AI applications \cite{back_clinical_cl}.




CL is a learning strategy that enables models to acquire a sequence of tasks. Such tasks may involve new skills, new data distributions, or different contexts. The primary aim of CL is to address the problem known as catastrophic forgetting \cite{back_survey_defying}. For AI models with fixed parameters, direct training on new tasks will lead to a significant performance drop on earlier tasks. 
A number of CL methods have been
proposed in recent years for various aspects of machine learning. 
By adopting regularization \cite{method_regu_ewc, method_regu_si}, extra memory that stores old samples \cite{method_replay_dgr, method_replay_gem}, or dynamic architectures \cite{method_dyn_wsn, method_dyn_pnn}, current CL methods are effective at retaining previously learned knowledge, and they have achieved promising results in classification problems \cite{back_survey_liyuan, back_survey_transai}.

\begin{figure*}[!ht]
    \centering
    \includegraphics[width=0.98\textwidth]{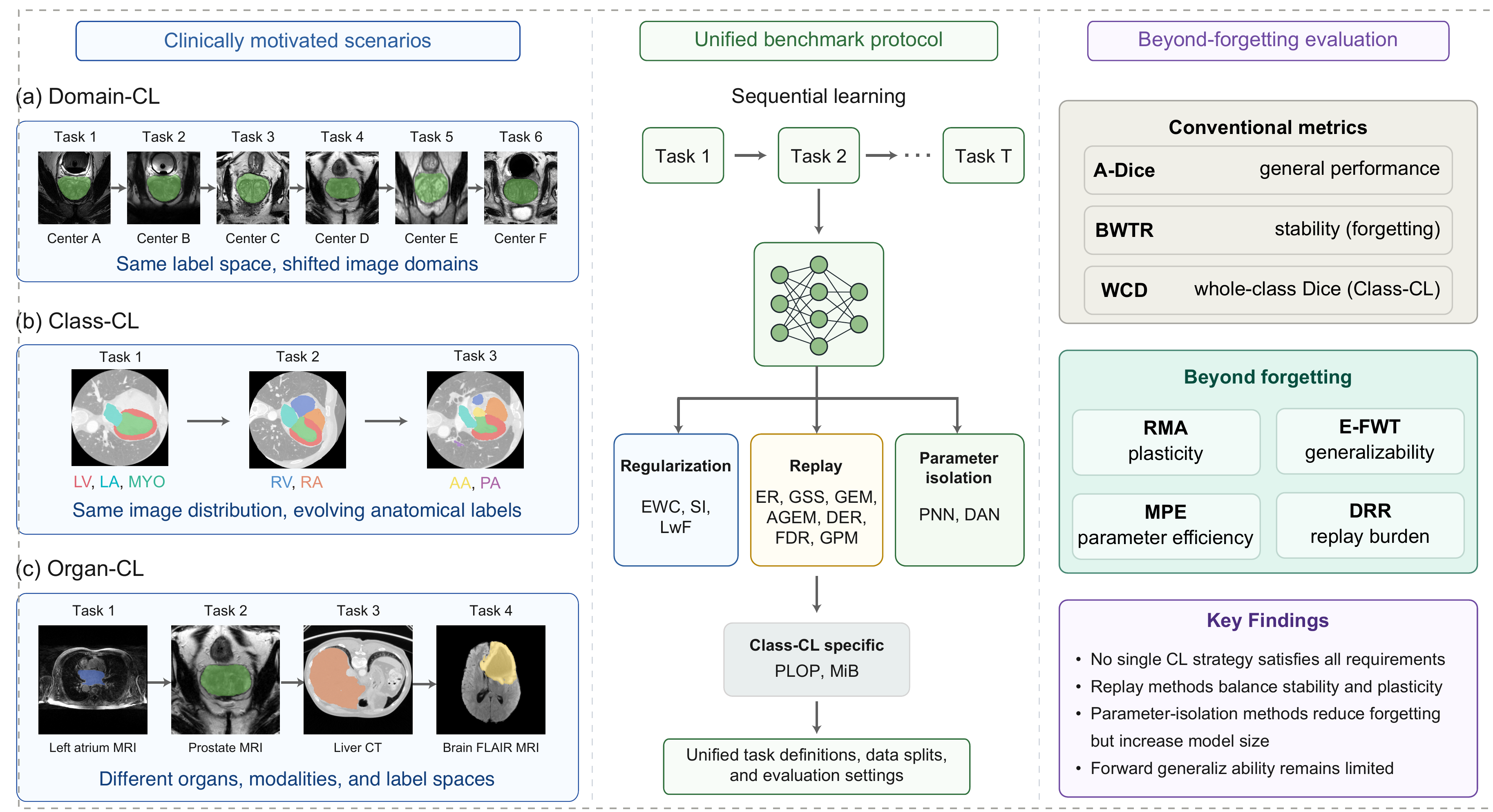}
    \caption{Overview of the proposed benchmark for continual medical image segmentation.}
    \label{fig:benchmark_overview}
\end{figure*}

Despite the critical need for CL in AI-driven medical applications, fundamental challenges remain due to the high-stakes nature, highly dynamic environments, and data privacy constraints. In clinical practice, CL may need to address a range of evolving conditions, such as data from new centers, the emergence of previously unseen diseases, and shifts in segmentation targets across diagnostic tasks. As a result, each scenario demands distinct modeling strategies and methodological considerations. To the best of our knowledge, no prior work has formally established a clear and unified definition of CL tailored to the medical image segmentation domain.
Besides, existing evaluation protocols mainly measure average performance and forgetting. These are insufficient for capturing the complex demands of high-stakes clinical tasks. The lack of a standardized benchmark dataset also limits the precise and comprehensive evaluation of CL methods. 
Moreover, current approaches emphasize mitigating forgetting, often neglecting model plasticity and generalizability, which are crucial for the real-world application of CL methods \cite{back_survey_defying}. Effective CL methods in medical image segmentation must balance competing objectives, including minimizing catastrophic forgetting, maintaining plasticity for new tasks, limiting access to past data due to privacy constraints of medical data, etc \cite{back_clinical_cl}. 
Currently, no single CL approach satisfies all these essential criteria simultaneously, highlighting an ongoing challenge in this research area.





In this paper, we present a benchmark study of CL for medical image segmentation. As summarized in Fig.~\ref{fig:benchmark_overview},
our benchmark is organized around three components:
clinically motivated continual segmentation scenarios, a unified
benchmark protocol for representative CL strategies, and an
evaluation framework that goes beyond forgetting. Our study comprehensively evaluates CL ability of representative approaches and examines critical properties for real-world deployment. To reflect real clinical settings, we define three CL scenarios that capture cross-center domain shift, incremental anatomical structure segmentation, and cross-organ segmentation. We further emphasize the stability–plasticity balance and introduce two metrics for its quantitative evaluation. Specifically, we introduce Restricted Modeling Ability (RMA) to measure model plasticity on new tasks, and extend the standard forward transfer metric (FWT) \cite{method_replay_gem} to Extended-FWT (E-FWT), which captures forward knowledge transfer across all unseen tasks. Finally, we conduct empirical evaluation of typical CL methods in our benchmark and discuss the future directions and challenges for CL in medical image segmentation.
To summarize, our contributions include:

\begin{itemize}

\item We present the desiderata of continual medical image segmentation in real clinical conditions and, accordingly, define three clinically inspired CL scenarios, targeting cross-center domain shifts, incremental anatomical structure segmentation, and cross-organ segmentation.

\item We establish an evaluation framework that goes beyond forgetting by assessing general performance, stability, plasticity, forward generalizability, parameter efficiency, and replay burden. In particular, we introduce RMA and E-FWT to quantify plasticity and forward transfer more comprehensively.

\item We evaluate the performance of representative CL methods in continual medical image segmentation, including regularization-based, replay-based, and architecture-based methods. In addition to key observations, we also discuss challenges and future directions for CL research in medical image segmentation.

\end{itemize}

\begin{table*}[!t]
\centering
\caption{Comparison of related benchmarks and surveys for continual medical image segmentation. ``D/C/O-CL'' denotes Domain-/Class-/Organ-CL scenarios. ``Stab.'', ``Plast.'', ``Gen.'', ``Param.'', and ``Repl.'' denote evaluation of stability, plasticity, forward generalizability, parameter efficiency, and replay burden, respectively.}
\label{tab:related_work}
\renewcommand{\arraystretch}{1.15}
\setlength{\tabcolsep}{5pt}
\begin{tabular*}{\textwidth}{@{\extracolsep{\fill}} l c c c c c c c c c c @{}}
\toprule
\multirow{2}{*}{Work} & \multirow{2}{*}{Type} & \multicolumn{3}{c}{Scenarios} & \multicolumn{5}{c}{Evaluation aspects} & \multirow{2}{*}{\makecell{Unified\\protocol}} \\
\cmidrule(lr){3-5} \cmidrule(lr){6-10}
 & & D-CL & C-CL & O-CL & Stab. & Plast. & Gen. & Param. & Repl. & \\
\midrule
\cite{clseg_survey}   & Survey (natural) & $\circ$ & $\circ$ & --      & \ding{55} & \ding{55} & \ding{55} & \ding{55} & \ding{55} & \ding{55} \\
\cite{medcl_survey}   & Survey (medical) & $\circ$ & $\circ$ & $\circ$ & \ding{55} & \ding{55} & \ding{55} & \ding{55} & \ding{55} & \ding{55} \\
\cite{lifelong_unet}  & Benchmark        & \ding{51} & \ding{55} & \ding{55} & \ding{51} & \ding{55} & \ding{55} & \ding{55} & \ding{55} & \ding{51} \\
\cite{whatwrong}      & Benchmark        & \ding{51} & \ding{55} & \ding{55} & \ding{51} & \ding{55} & \ding{55} & \ding{55} & \ding{55} & \ding{51} \\
\textbf{Ours}         & \textbf{Benchmark} & \textbf{\ding{51}} & \textbf{\ding{51}} & \textbf{\ding{51}} & \textbf{\ding{51}} & \textbf{\ding{51}} & \textbf{\ding{51}} & \textbf{\ding{51}} & \textbf{\ding{51}} & \textbf{\ding{51}} \\
\bottomrule
\end{tabular*}
\vspace{2pt}
\footnotesize
\raggedright
$\circ$: discussed conceptually but not benchmarked. \ding{51}: covered with experiments / metrics. \ding{55}: not covered. --: out of scope.
\end{table*}

\section{Related Work}

Prior work on continual learning for medical image segmentation has mainly taken two forms: surveys that summarize methodological progress and benchmarks that empirically compare existing methods. We review both lines of work and clarify the gap that motivates our benchmark.

\subsection{Related Surveys on Continual Learning}

Broad surveys of continual learning~\cite{general_survey} have established the standard taxonomy of regularization-based, replay-based, and parameter-isolation methods. These surveys have provided an important foundation for the field. However, they are largely built around classification benchmarks, where the prediction target is a single image-level label. Their evaluation protocols therefore focus mainly on average accuracy and backward transfer. Such metrics are insufficient for medical image segmentation, where models must preserve dense anatomical predictions, adapt to evolving label spaces, and remain practical under privacy and resource constraints.

Continual semantic segmentation has also been reviewed in the context of dense prediction~\cite{clseg_survey}. This line of work identifies important challenges, including the stability-plasticity trade-off, semantic drift, and background shift. Nevertheless, existing analyses are primarily based on natural-image datasets, such as PASCAL VOC and ADE20K. These settings differ substantially from medical imaging. Cross-center domain shifts may change image appearance while preserving the label space. Incremental anatomical segmentation introduces background shift under strong anatomical and spatial priors. Cross-organ segmentation further requires learning tasks with disjoint label spaces and different image distributions. These clinically relevant settings are not fully captured by natural-image continual segmentation.

A recent survey on continual learning in medical image analysis~\cite{medcl_survey} provides a broader overview across classification, detection, and segmentation. While valuable, its scope does not allow a dedicated treatment of continual medical image segmentation. In particular, it does not provide a clinically grounded scenario taxonomy, a unified evaluation protocol beyond forgetting, or a systematic empirical comparison of representative methods under common settings. Our work addresses this gap by formalizing three continual segmentation scenarios and evaluating representative continual learning strategies with metrics that jointly assess stability, plasticity, generalizability, parameter efficiency, and replay burden.

\subsection{Benchmarks for Continual Medical Segmentation}

Only a few studies have attempted to establish standardized benchmarks for continual medical image segmentation. Lifelong nnU-Net~\cite{lifelong_unet} represents an important step toward reproducible evaluation by providing a sequential training and testing framework built on nnU-Net. However, it mainly focuses on cross-center domain continual learning, evaluates a limited set of methods, and relies primarily on Dice and backward transfer. As a result, several deployment-relevant properties remain unmeasured, including plasticity, forward generalizability, parameter growth, and replay burden.

Gonzalez et al.~\cite{whatwrong} further questioned whether existing continual learning methods are sufficiently competitive against strong multi-model baselines, and proposed UNEG as a practical alternative based on separate segmenters and autoencoders. This study provides an important critique of the added complexity of continual learning methods. Yet its evaluation is still centered on a domain-shift setting and a relatively narrow set of metrics. It therefore leaves open how continual learning methods behave under class-incremental and organ-incremental segmentation, both of which are common in evolving clinical workflows.

Table~\ref{tab:related_work} summarizes the comparison between our benchmark and closely related surveys and benchmarks. The comparison highlights that existing works either provide broad methodological reviews or focus mainly on Domain-CL, whereas our benchmark explicitly covers Domain-CL, Class-CL, and Organ-CL and evaluates deployment-relevant properties beyond forgetting, including parameter efficiency and replay budget. Our benchmark differs from these studies in three main respects. First, we define three clinically motivated scenarios, namely Domain-CL, Class-CL, and Organ-CL, which capture different forms of clinical evolution rather than a single domain-shift setting. Second, we introduce an evaluation framework that goes beyond forgetting by jointly measuring plasticity, forward generalizability, parameter efficiency, and replay burden together with conventional performance and stability metrics. Third, we evaluate representative methods from regularization-based, replay-based, and parameter-isolation families under unified task definitions, data splits, and evaluation protocols. We further complement the main benchmark with task-order robustness and buffer-size analyses. By keeping data, splits, and metrics consistent across methods, our benchmark reveals recurrent trade-offs that are difficult to identify from method-specific studies alone. For example, the results in Section~\ref{sec_result} show that current strategies remain limited in forward generalizability, even when they reduce forgetting effectively. These findings provide a more systematic basis for developing clinically deployable continual medical image segmentation systems.

\section{Definitions and Materials} 

This section first defines three CL scenarios for medical image segmentation in Section~\ref{sec:2.1}. 
Then, we discuss the desiderata of continual medical image segmentation, in Section~\ref{sec:2.2}, which motivates us to form a set of  metrics for comprehensive evaluation of benchmarked methods, in Section~\ref{sec:2.3}. 
Finally, Section~\ref{sec:2.4} and \ref{sec:2.5} list the methods and datasets used in the benchmark studies.

\begin{table*}[!t]
  \centering
  \caption{The three continual medical image segmentation scenarios.}
  \label{tab:CL_settings}
  \renewcommand{\arraystretch}{1.25}
  \setlength{\tabcolsep}{10pt}
  \begin{tabular*}{0.7\textwidth}{@{\extracolsep{\fill}} l c c c @{}}
    \toprule
    \textbf{Scenario} & \textbf{Condition} & \textbf{Label Space} & \textbf{Image Distribution} \\
    \midrule
    Domain-CL  
      & \multirow{3}{*}{$\forall t_1,t_2\in\mathcal{T}$ and $t_1 \neq t_2$}
      & $Y_{t_1} = Y_{t_2}$ 
      & $P(\mathcal{X}_{t_1}) \neq P(\mathcal{X}_{t_2})$ \\
    Class-CL    
      & 
      & $Y_{t_1} \neq Y_{t_2}$ 
      & $P(\mathcal{X}_{t_1}) = P(\mathcal{X}_{t_2})$ \\
    Organ-CL   
      & 
      & $Y_{t_1} \cap Y_{t_2} = \emptyset$
      & $P(\mathcal{X}_{t_1}) \neq P(\mathcal{X}_{t_2})$ \\
    \bottomrule
  \end{tabular*}
\end{table*}

\subsection{Three scenarios of continual medical image segmentation} \label{sec:2.1}

We study three CL scenarios for medical image segmentation, \ie domain continual learning (Domain-CL), class continual learning (Class-CL), and organ continual learning (Organ-CL).

Let $t$ be the timestep index of a learning task denoted by $f_t$, and $t\in \mathcal{T} = \{1, 2, \dots, T\}$.   
The training dataset of $f_t$ is given by $\mathcal{D}_t = \{(x_i^t, y_i^t)\}_{i=1}^{N_t}$, where $x_i^t$ and $y_i^t$ are respectively the image and segmentation of the $i^{th}$ training sample, and $N_t$ is the number of samples. 
We assume that training samples are drawn i.i.d. from the joint distribution, \ie $\mathbb{D}_t = P(\mathcal{X}_t, \mathcal{Y}_t)$, where $\mathcal{X}_t$ and $\mathcal{Y}_t$ are the random variables of image and segmentation, respectively.  
For segmentation task $f_t$, we define the set of all label classes using $Y_t$, of which each represents an anatomical structure or substructure in medical images.
Table~\ref{tab:CL_settings} provides definitions of the three scenarios, which are elaborated on in the following.

\renewcommand{\arraystretch}{1.3} 

\renewcommand{\arraystretch}{1.0} 



\begin{itemize}

\item\textbf{Domain-CL} aims to address the challenge of learning from a series of domains that exhibit distributional shifts. 
This is because medical images are typically generated in decentralized clinical centers. It is difficult to aggregate data across centers, due to privacy concerns. 
Hence, we propose to use CL methods to learn the domains sequentially, \ie  to learn continually.
In such a scenario, the segmentation label space is consistent across domains, i.e., $Y_{t_1} = Y_{t_2}$,  for $\forall t_1,t_2\in\mathcal{T}$ and $t_1 \neq t_2$, but the image distributions vary, \ie $P(\mathcal{X}_{t_1}) \neq P(\mathcal{X}_{t_2})$. 
The distribution shift is due to various factors, including differences in scanners, imaging protocols and patient demographics. 


\item 
\textbf{Class-CL} focuses on multi-class segmentation tasks, where a model incrementally learns to segment new anatomical (sub)-structures within the same fields of view (FOV) or within an organ. 
In such scenario, the model learns new labels in each task, but the image distribution remains the same, \ie 
for $\forall t_1,t_2\in\mathcal{T}$,  we have $Y_{t_1} \neq Y_{t_2}$ and $P(\mathcal{X}_{t_1}) = P(\mathcal{X}_{t_2})$.  

Given cardiac segmentation as an example, the myocardial viability study may be interested in segmenting the two ventricular cavities and left ventricle myocardium from cardiac CT image, as the initial task. 
By contrast, the atrial fibrillation group may need the AI model to continually learn the segmentation of atrial substructures, as a followup task. 
Finally, cardiologists specialized in Transcatheter Aortic Valve  Replacement (TAVR) may require the model to further include the segmentation of aorta for visualizing the whole heart structure in the surgical plan of TAVR, which is then defined as the third task in the Class-CL scenario.

\item \textbf{Organ-CL} involves continual segmentation of multiple organs and associated pathological structures (e.g., tumors).
In this scenario, the image content differs across tasks, and thus the label space varies. 
Hence, for two different tasks $t_1$ and $t_2$ Organ-CL assumes $P(\mathcal{X}_{t_1}) \neq P(\mathcal{X}_{t_2})$ and $Y_{t_1} \cap Y_{t_2} = \emptyset$.
Given metastatic cancer as an example,  continual segmentation of primary and metastatic organs with their associated lesions is essential for tracking tumor spread and guiding therapy. 
\end{itemize}

\subsection{Desiderata of continual learning: Beyond forgetting} \label{sec:2.2}


While mitigating catastrophic forgetting remains a major concern in CL research, real-world clinical applications demand more clinically relevant capabilities. We identify the following key desiderata for continual medical image segmentation. 
 

\begin{itemize}
\item \textbf{Stability.} 
It refers to a model's capacity to retain previously acquired \emph{old} knowledge while learning \emph{new} tasks.
In the standard static training paradigm, AI models frequently exhibit catastrophic forgetting -- a phenomenon where prior knowledge is overwritten during incremental learning.   
Hence, ensuring stability of performance across sequential tasks remains a major challenge in CL research.  
    
\item \textbf{Plasticity.}
It denotes the capacity to acquire knowledge efficiently from novel tasks.
Retaining such adaptability is important in CL as additional tasks are introduced. 
However, in practice a trade-off emerges:

- Strengthened stability can constrain plasticity of a model, hindering its adaptation to new tasks.

- Conversely, enhanced plasticity often exacerbates catastrophic forgetting by disrupting previously learned representations.

Hence, maintaining \textbf{stability-plasticity balance} should be a core objective in CL research  to sustain a robust general performance across all tasks.

\item \textbf{Minimal replay.}  
CL must optimize for minimal data replay from prior tasks, particularly in medical imaging where stringent privacy regulations (e.g., HIPAA \cite{replay_HIPAA} in the U.S./ GDPR \cite{replay_GDPR} in Europe) prohibit the retention or reuse of patient data.  
Furthermore, these methods should operate under strict memory constraints by avoiding extensive storage of auxiliary data that would otherwise incur significant memory overhead.
Such auxiliary data can be  computational artifacts, such as intermediate feature maps or gradient space from previous tasks.

\item \textbf{Low parameter growth.} 
CL must achieve task acquisition with minimal expansion of model parameters. 
This constraint is essential for maintaining computational efficiency during both training and deployment phases, particularly in clinical environments where computational resources and run time are constrained.   
\end{itemize}

Beyond these fundamental capabilities, effective deployment of CL systems could necessitate more critical attributes, including cross-domain generalizability, task-order robustness, and task-agnostic learning ability. 
  \textbf{Cross-domain generalizability} demands effective knowledge transfer to novel tasks and out-of-distribution data scenarios.
  \textbf{Task-order robustness} requires model performance to remain invariant to permutations in task sequences, ensuring consistent behavior regardless of presentation order. 
  \textbf{Task-agnostic learning} implies that neither training nor inference processes should depend on predefined task identity or explicit boundaries.
In summary, these properties determine the clinical viability of CL systems, where adherence to these criteria directly translates to enhanced real-world applicability.

\subsection{Evaluation metrics} \label{sec:2.3}  

We comprehensively evaluate continual segmentation methods from five aspects. 
The fundamental metric is segmentation accuracy, for which we use the widely adopted Dice score.
Let $d_{t,i}$ be the average Dice of a model trained after the $t$-th task and tested on the data of the \(i\)-th task.   
The five aspects are as follows.

\begin{itemize}
\item \textbf{General performance across CL tasks.}   
We measure the average segmentation accuracy on all learning tasks as the general performance of a CL method.
Hence, the general performance, referred to as A-Dice, is computed from averaged Dice, as follows, 
\begin{equation}
\mbox{A-Dice}=\frac{1}{T}\sum_{i=1}^T{d_{T,i}} .
\end{equation}
\item \textbf{Model stability.} Backward Transfer (BWT) is widely used to measure stability \cite{method_replay_gem}. However, BWT is sensitive to the absolute performance scale, which limits fair comparison across different tasks. 
Here, we propose \textit{BWT Ratio (BWTR)},
\begin{equation}
\mbox{BWTR}=\frac{1}{T-1}\sum_{i=1}^{T-1}\frac{d_{T,i}-d_{i,i}}{d_{i,i}} ,
\end{equation}

\item \textbf{Model plasticity.} 
We propose a new metric, \ie the Restricted Modeling Ability (RMA), to assess the learning plasticity of a model, 
\begin{equation}
\mbox{RMA}=\frac{1}{T-1}\sum_{i=2}^T{\frac{d_{i,i}}{d^{NC}_i}},
\end{equation}
where, $d^{NC}_i$ denotes the Dice accuracy of a \emph{non-CL} model solely trained using the training data of the $i$-th task and tested on the $i$-th testset, representing the plasticity upperbound. \emph{RMA is a critical metric indicating segmentation performance on new tasks}.
An RMA value close to 1 indicates perfect plasticity, suggesting the model continually learns new tasks with the same ability as independent training without any constraint from CL. 
Conversely, RMA values significantly lower than 1 suggest impaired plasticity.

\item \textbf{Model generalizability.} 
For Domain-CL, we propose the Extended Forward Transfer (E-FWT) metric to evaluate model generalizability. \emph{Note that generalization is specific to Domain-CL, as in Class-CL and Organ-CL the segmentation targets differ across tasks, making cross-task generalization not directly applicable.} The conventional Forward Transfer (FWT) metric evaluates generalization performance solely on the immediate one follow-up task \cite{method_replay_gem}, making it highly sensitive to task order and discrepancies between adjacent tasks.  
The proposed E-FWT metric is defined as follows,
\begin{equation}
\mbox{E-FWT}=\frac{2}{T(T-1)}\sum_{t=1}^{T-1}\sum_{i=t+1}^{T}
({d_{t,i}}-d_{0,i})   ,
\end{equation}
where $d_{0,i}$ denotes the Dice accuracy of a randomly initialized model without any training, evaluated on the test data of the $i$-th task. Note that some works use FWT to measure plasticity \cite{back_survey_liyuan, method_replay_gem}. We distinguish between model generalization and plasticity, and use the E-FWT and RMA metrics to quantify them separately.

\item \textbf{Whole class segmentation performance.} 
For Class-CL, we report the segmentation performance over all classes encountered after completing the final task. 
We denote this metric as Whole Class Dice (WCD), defined as $\text{WCD} = d_{\text{all}}$, where $d_{\text{all}}$ denotes the Dice score of all seen classes.
\end{itemize}

In addition to the above CL task performance metrics, we evaluate the model parameter efficiency and volume of replay data using the following metrics,
\begin{itemize}
\item \textbf{Model parameter efficiency.} 
We use the model parameter efficiency (MPE) to evaluate model parameter growth:
\begin{equation}
    \mbox{MPE} = \frac{1}{T-1}
    \sum_{t=2}^{T} \frac{\mathrm{Param}(\theta_{t})-\mathrm{Param}(\theta_{t-1})}{\mathrm{Param}(\theta_1)},
\end{equation}
where $\mathrm{Param}(\theta_t)$ denotes the number of  model parameters at task $t$.
Note that MPE measures the parameter growth rate rather than the absolute number of parameters, as all CL methods in this study are initialized with the same backbone network. 

\item \textbf{Data replay ratio.}  
We first define replay information as the additional content apart from the trained model that is brought to the next task.
Since replaying original medical images can pose significant privacy concerns in clinical practice, we use a binary metric to indicate whether a method replays raw images. 
We then define the Data Replay Ratio (DRR) to quantify the amount of past-task data used for rehearsal:  
\begin{equation}
    \mbox{DRR} = \frac{1}{T-1}\sum_{t=1}^{T-1} \frac{N^{\mathrm{replay}}_{t}}{ N^{\mathrm{train}}_{t}},
\end{equation}
where $N^{\mathrm{replay}}_{t}$ denotes the number of training samples from task $t$ either directly used in follow-up tasks or used to generate replay information, and $N^{\mathrm{train}}_{t}$ denotes the number of training samples of task $t$.  
Note that such replay information can be contents trained or distilled from samples of previous tasks, such as feature representations or gradient space.
\end{itemize}

\subsection{Benchmark methods} \label{sec:2.4} 

In our benchmark, we evaluate representative continual learning methods. For medical image segmentation, three CL strategies can be adopted, \ie (1) regularization, (2) replay and (3) parameter-isolation, which are described in detail in Section~\ref{regu}, \ref{repl} and \ref{ParIso}, respectively. 

In addition, for the Class-CL scenario we implement two representative methods for comparisons, \ie Pseudolabel and local pooled outputs distillation (referred to as ClassCL-PLOP) \cite{method_seg_plop} and Modeling the Background (denoted as ClassCL-MiB) \cite{method_seg_mib}:
\begin{itemize}
    \item \textbf{ClassCL-PLOP} uses multi-scale pooling distillation and entropy-based pseudo-labeling of the background to reduce forgetting of old classes.
 \item \textbf{ClassCL-MiB} mitigates forgetting by redefining the background label and adapting the distillation loss to account for both old and new classes. 
\end{itemize}

Finally, we compare the CL methods with the non-continual baseline (referred to as Non-CL) and joint training scheme (denoted as JointTrain): 
\begin{itemize}
    \item \textbf{Non-CL} adopts the vanilla training scheme without using any CL strategy in new tasks, 
thus its segmentation accuracy represents a lower bound performance for CL methods.
    \item \textbf{JointTrain} centralizes the available data from all tasks, and trains the model using all the training image and segmentation labels simultaneously. 
This method is similar to a method adopting the replay strategy with 100\% access to the whole training datasets of previous tasks, thus its accuracy can be considered as an upper bound for CL methods.
\end{itemize}

\subsubsection{Regularization strategy} \label{regu}
The regularization strategy introduces explicit terms to the loss function. This strategy includes \textbf{weight regularization} and \textbf{function regularization}. Weight regularization constrains changes in network parameters.
Function regularization constrains intermediate or final outputs with knowledge distillation. In this work, we benchmark the well-known methods adopting the regularization strategy, 
including, 
\begin{itemize}
\item Regu-EWC: Elastic Weight Consolidation (EWC) estimates the Fisher information matrix to regularize changes of important weights \cite{method_regu_ewc}. This is a weight regularization strategy.
\item Regu-SI: Synaptic Intelligence (SI) evaluates the parameter importance according to the relative weight changes in the optimization process \cite{method_regu_si}. Regu-SI belongs to the weight regularization strategy.
\item Regu-LwF: Learning Without Forgetting (LwF) preserves previous knowledge by computing a distillation loss using new data and their predictions from the output head of the old tasks\cite{method_regu_lwf}. We choose Regu-LwF as a representative function regularization strategy. 
 
\end{itemize}

\subsubsection{Replay strategy} \label{repl}
This strategy mitigates forgetting by including a small number of original data or contents that were trained or distilled from original images of previous tasks. The representative methods include, 
\begin{itemize}
\item Repl-ER: Experience Replay (ER)  directly samples past data from the memory buffer and combines it with new task data for training the new task \cite{method_replay_er}.
\item Repl-GSS: Gradient-based Sample Selection (GSS) improves the sample diversity of the memory buffer with parameter gradients \cite{method_replay_gss}. 
\item Repl-GEM: Gradient Episodic Memory (GEM) constructs individual constraints based on the old training samples to ensure no increase in their losses \cite{method_replay_gem}. 
\item Repl-GPM: Gradient Projection Memory (GPM) projects new task gradient orthogonal to the important gradient subspace of previous tasks \cite{method_regu_gpm}.
\item Repl-AGEM: Averaged Gradient Episodic Memory (AGEM) further improves training efficiency by replacing task-specific constraints with a single global loss across all tasks \cite{method_replay_agem}.
\item Repl-DER: Dark Experience Replay (DER) retains the logits of past samples and leverages knowledge distillation to mitigate forgetting, while DER++ further stores both the raw data and their corresponding logits in the buffer to enhance CL performance\cite{method_replay_der}. 
\item Repl-FDR: Function Distance Regularization (FDR) constrains changes in the input-output mapping of previous tasks by computing function distance on memory data \cite{method_replay_fdr}.
\end{itemize}

Except for Repl-GPM, dependence on storing and replaying raw samples from earlier tasks is a notable challenge of these methods. It could raise considerable privacy concerns, particularly in practical clinical applications. 

\subsubsection{Parameter-isolation strategy} \label{ParIso}
The parameter-isolation strategy constructs task-specific parameters for each task, promoting learning capacity as well as preventing forgetting, at the cost of increasing model size. The benchmarked methods are as follows,
\begin{itemize}
\item ParIso-PNN: Progressive Neural Networks (PNN) introduces a specific module for each task, and enables knowledge transfer through adaptor connections between these sub-models \cite{method_dyn_pnn}.
\item ParIso-DAN: Deep Adaptation Network (DAN) introduces controller modules to linearly combine newly added parameters and old parameters \cite{method_dyn_dan}. 
\end{itemize}

\subsection{Benchmark datasets} \label{sec:2.5}
 
We benchmarked the three continual medical image segmentation scenarios using a variety of datasets. 
For each segmentation task, we randomly split them into 3 sets, with a set of $60\%$ cases for training, a set of $15\%$   for validation, and a set of $25\%$   for test. 
Note that the validation set was used for selection of hyper-parameters.
 
For \textbf{Domain-CL} studies, we constructed a six-center prostate T2 MRI dataset from three publicly available datasets, \ie NCI-ISBI13 (Centers A and B) \cite{dataset_prostate2},  I2CVB (Center C) \cite{dataset_prostate}, and PROMISE12 (Center D, E and F) \cite{dataset_prostate3}.  
Images from the six centers, representing 6 domains, demonstrate diverse scanning protocols and demographics \cite{dataset_domaincl}.  
For preprocessing, we cropped all the images from Center C, to ensure that the axial views were aligned with images from the other centers. 
We then resized all images to be $256\times256$ in the axial plane, and normalized the intensity range per case using the mean and standard deviation of non-zero voxels.


For \textbf{Class-CL} studies, we adopted cardiac CT from the Multi-Modality Whole Heart Segmentation Challenge (MMWHS) \cite{care25ASA1, discussion_inter_seg3, care25ASA2}. 
MMWHS provides manual segmentation of seven whole heart structures. The slices were acquired in the axial view. The in-plane resolution is \(0.78 \times 0.78\) mm, and the average slice thickness is 1.60 mm. We resized all slices to be $256\times256$.
In the Class-CL setting, we split the segmentation classes into three tasks according to heart structures. Specifically, the first task includes the Left Ventricle (LV), Left Atrial (LA), and Myocardium (MYO). The second segmentation task consists of the Right Ventricle (RV) and Right Atrium (RA). The Ascending Aorta (AA) and Pulmonary Artery (PA) are included in the third segmentation task. 

For \textbf{Organ-CL} studies, we considered four publicly available datasets with different anatomical structures and imaging modalities. Specifically, we selected cardiac LGE MRI from Left Atrial and Scar Quantification and Segmentation Challenge (LAScarQS) \cite{dataset_lascars, dataset_lascars2}, prostate T2 MRI from PROMISE12 (Center D in Domain-CL) \cite{dataset_prostate3}, liver CT from The Liver Tumor Segmentation Challenge (LiTs) \cite{dataset_taskcl}, and brain FLAIR MRI from Federated Tumor Segmentation Challenge 2021(FeTS) \cite{dataset_taskcl2}. The Organ-CL tasks were arranged sequentially as follows: (1) left atrium segmentation, (2) prostate segmentation, (3) liver segmentation, and (4) brain tumor segmentation. Each image was normalized to zero mean and unit variance, and slices were resized to $256\times256$.

\section{Results}
\label{sec_result}

This section first studies the benchmarking methods on the three scenarios, \ie Domain-CL, Class-CL, and Organ-CL, and analyzes their CL performance in Section~\ref{sec:3.1}, \ref{sec:3.2}, and \ref{sec:3.3}, respectively. 
Section~\ref{sec:3.4} then presents the results of task-ordering experiments, and Section~\ref{sec:3.5} investigates the influence of buffer size on replay-based methods.
 
\subsection{Domain-CL Study} \label{sec:3.1}

In this study, we analyze the performance of representative CL strategies in the Domain-CL setting based on the evaluation metrics defined in Section \ref{sec:2.3}. Note that the RMA value of Non-CL is 1.086, which indicates that continual training on sequential and similar prostate datasets improves model plasticity in this specific setting.

\textbf{Regularization-based methods demonstrated suboptimal general performance, and weight regularization strategies struggled to mitigate catastrophic forgetting.}
Among the three regularization-based methods (Regu-EWC, Regu-SI, Regu-LwF), the average A-Dice was only 0.670, which is lower than the replay-based average of 0.729 and substantially below the JointTrain upper bound of 0.830. These results indicate inferior general segmentation performance in domain-CL segmentation tasks. Regarding stability, both Regu-EWC and Regu-SI exhibited significant forgetting, with BWTR values of –0.242 and –0.236, respectively. They are close to the Non-CL lower bound of –0.196. In contrast, Regu-LwF achieved high stability, with a BWTR of $-0.090$.
These findings suggest that parameter-level regularization alone is insufficient to prevent forgetting. Alternatively, function-level regularization helps the model preserve its overall response behavior, thereby better retaining previously learned knowledge.
However, Regu-LwF obtained the lowest RMA (0.894) among all evaluated methods, implying that its improved stability may come at the cost of reduced model plasticity.

\textbf{Replay-based methods generally achieved favorable stability while maintaining model plasticity.}
As Table \ref{tab:domain-cl} shows, the majority of replay-based methods achieved strong general performance. For instance, Repl-ER obtained an A-Dice of 0.750, and Repl-AGEM reached 0.740. These methods also maintained relatively low forgetting, reflected in BWTR values close to zero, such as –0.100 for Repl-ER and –0.105 for Repl-DER++. Furthermore, Repl-ER and Repl-GSS demonstrated high RMA scores, indicating preserved plasticity. Specifically, Repl-ER reached 1.054, while Repl-GSS achieved 1.059. These outcomes show that replay-based strategies are able to achieve a favorable stability–plasticity trade-off, enabling both effective knowledge retention and adaptation to new tasks. 
In our implementation, we fixed the buffer size at 32 for all methods. The variation in DRR values among replay-based methods arises from differences in how each method populates and updates its buffer. The results indicate that even with an average raw data replay rate as low as 2.7\% or 4.2\%, these methods can deliver competitive segmentation performance.

\textbf{Parameter-isolation strategies eliminated forgetting while achieving high general performance.}
As Table~\ref{tab:domain-cl} shows, ParIso-PNN and ParIso-DAN achieved the highest A-Dice scores, 0.789 and 0.786 respectively, with zero forgetting in all Domain-CL tasks. This was achieved through allocating an independent sub-network to each task while facilitating cross-task knowledge transfer via adaptor connections. However, these methods need to maintain a separate parameter set for each task, resulting in continual model parameter growth. This is reflected in the high MPE values of $1.112$ and $0.111$ for the two methods, respectively. Notably, ParIso-DAN achieved comparable performance to ParIso-PNN with a substantially lower parameter growth rate, indicating more efficient use of additional parameters.

\textbf{Generalizability is still a persistent challenge across all CL strategies.} As Table \ref{tab:domain-cl} shows, most CL methods achieve E-FWT scores below the Non-CL baseline (0.260), indicating that current strategies generally fail to enhance forward transfer over naive sequential training. This arises from the lack of mechanisms explicitly designed to enhance forward transfer in current CL research.
As a result, generalizability to unseen domains remains a critical yet unresolved challenge, underscoring the need for strategies that promote forward knowledge transfer to improve segmentation performance across diverse imaging protocols and clinical centers.


\begin{table*}[]
\centering
\caption{Results of Domain-CL. An asterisk in the DRR column indicates that the method replays raw samples from previous tasks.}
\label{tab:domain-cl}
\renewcommand{\arraystretch}{1.2}
\begin{small}
\begin{tabular}{@{}ccccccc@{}}\toprule
\textbf{Method} & \textbf{A-Dice $\uparrow$} & \textbf{BWTR $\uparrow$} & \textbf{RMA $\uparrow$} & \textbf{E-FWT $\uparrow$} & \textbf{MPE $\downarrow$} & \textbf{DRR $\downarrow$}\\ \midrule
Regu-EWC & $.672\pm .021$ & $-0.242\pm .039$ & $1.086\pm .014$ & $.255\pm .152$ & $0$ & $0$ \\ \hdashline
Regu-SI & $.676\pm .024$ & $-0.236\pm .042$ & $1.084\pm .008$ & $.259\pm .151$ & $0$ & $0$ \\ \hdashline
Regu-LwF & $.663\pm .029$ & $-0.090\pm .027$ & $.894\pm .021$ & $.224\pm .153$ & $0$ & $0$ \\ \midrule

Repl-ER & $.750\pm .021$ & $-0.100\pm .027$ & $1.054\pm .014$ & $.256\pm .150$ & $0$ & $.027^{*}$ \\ \hdashline
Repl-GSS & $.726\pm .026$ & $-0.140\pm .040$ & $1.059\pm .010$ & $.255\pm .150$ & $0$ & $.036^{*}$ \\ \hdashline
Repl-GEM & $.718\pm .018$ & $-0.095\pm .022$ & $.994\pm .024$ & $.251\pm .154$ & $0$ & $.042^{*}$ \\ \hdashline
Repl-GPM & $.661\pm .010$ & $-0.188\pm .010$ & $1.013\pm .022$ & $.212\pm .080$ & $0$ & $.113$ \\ \hdashline
Repl-AGEM & $.740\pm .015$ & $-0.093\pm .023$ & $1.023\pm .027$ & $.253\pm .155$ & $0$ & $.042^{*}$ \\ \hdashline
Repl-DER & $.715\pm .011$ & $-0.126\pm .015$ & $1.018\pm .016$ & $.249\pm .148$ & $0$ & $.027^{*}$ \\ \hdashline
Repl-DER++ & $.725\pm .013$ & $-0.105\pm .018$ & $1.012\pm .013$ & $.251\pm .149$ & $0$ & $.027^{*}$ \\ \hdashline
Repl-FDR & $.728\pm .015$ & $-0.111\pm .028$ & $1.025\pm .033$ & $.252\pm .136$ & $0$ & $.042^{*}$ \\ \midrule

ParIso-PNN & $.789\pm .002$ & $0\pm 0$ & $1.003\pm .014$ & $.235\pm .148$ & $1.112$ & $0$ \\ \hdashline
ParIso-DAN & $.786\pm .008$ & $0\pm 0$ & $1.016\pm .021$ & $.312\pm .210$ & $.111$ & $0$ \\ \midrule

Non-CL & $.676\pm .021$ & $-0.237\pm .038$ & $1.086\pm .008$ & $.260\pm .151$ & $0$ & $0$ \\ \midrule
JointTrain & $.830\pm .040$ & / & / & / & $0$ & $0$ \\
\bottomrule
\end{tabular}
\end{small}
\end{table*}

\subsection{Class-CL Study} \label{sec:3.2}

\begin{table*}[t]
\centering
\caption{Results of Class-CL. An asterisk in the DRR column indicates that the method replays raw samples from previous tasks.}
\label{tab:class-cl}
\renewcommand{\arraystretch}{1.2}
\setlength{\tabcolsep}{4pt}
\begin{tabular}{@{}ccccccc@{}}\toprule
\textbf{Method} & \textbf{A-Dice $\uparrow$} & \textbf{BWTR $\uparrow$} & \textbf{RMA $\uparrow$} & \textbf{WCD $\uparrow$} & \textbf{MPE $\downarrow$} & \textbf{DRR $\downarrow$} \\ \midrule
Regu-EWC & $.487\pm .020$ & $-0.552\pm .042$ & $.999\pm .022$ & $.513\pm .022$ & $0$ & $0$ \\ \hdashline
Regu-SI & $.469\pm .007$ & $-0.602\pm .016$ & $1.018\pm .028$ & $.570\pm .020$ & $0$ & $0$ \\ \hdashline
Regu-LWF & $.384\pm .022$ & $-0.575\pm .013$ & $.782\pm .046$ & $.602\pm .002$ & $0$ & $0$ \\ 
\midrule

Repl-ER & $.654\pm .034$ & $-0.243\pm .068$ & $1.015\pm .029$ & $.596\pm .019$ & $0$ & $.011^{*}$ \\ \hdashline
Repl-GSS & $.456\pm .003$ & $-0.611\pm .016$ & $.991\pm .038$ & $.600\pm .006$ & $0$ & $.018^{*}$ \\ \hdashline
Repl-GEM & $.684\pm .036$ & $-0.304\pm .038$ & $1.014\pm .031$ & $.622\pm .031$ & $0$ & $.010^{*}$ \\ \hdashline
Repl-GPM & $.575\pm .023$ & $-0.219\pm .033$ & $.892\pm .014$ & $.513\pm .014$ & $0$ & $.068$ \\ \hdashline
Repl-AGEM & $.595\pm .017$ & $-0.370\pm .032$ & $1.020\pm .019$ & $.562\pm .026$ & $0$ & $.010^{*}$ \\ \hdashline
Repl-DER & $.628\pm .027$ & $-0.219\pm .070$ & $.913\pm .029$ & $.622\pm .029$ & $0$ & $.011^{*}$ \\ \hdashline
Repl-DER++ & $.638\pm .036$ & $-0.208\pm .056$ & $.932\pm .034$ & $.633\pm .043$ & $0$ & $.011^{*}$ \\ \hdashline
Repl-FDR & $.666\pm .012$ & $-0.146\pm .046$ & $.917\pm .035$ & $.627\pm .015$ & $0$ & $.010^{*}$ \\ \midrule

ParIso-PNN & $.770\pm .012$ & $0\pm 0$ & $.976\pm .021$ & $.620\pm .020$ & $1.056$ & $0$ \\ \hdashline
ParIso-DAN & $.729\pm .009$ & $0\pm 0$ & $.897\pm .022$ & $.543\pm .127$ & $.111$ & $0$ \\ \midrule

ClassCL-PLOP & $.708\pm .025$ & $-0.104\pm .026$ & $.986\pm .038$ & $.707\pm .009$ & $0$ & $0$ \\ \hdashline
ClassCL-MiB & $.751\pm .008$ & $-0.039\pm .003$ & $1.004\pm .018$ & $.697\pm .008$ & $0$ & $0$ \\ \midrule
Non-CL & $.471\pm .012$ & $-0.598\pm .015$ & $1.014\pm .028$ & $.572\pm .023$ & $0$ & $0$ \\ \midrule
JointTrain & $0.810 \pm 0.023$ & / & / & / & $0$ & $0$ \\
\bottomrule
\end{tabular}
\end{table*}

In Class-CL segmentation, forgetting introduced by background shift becomes a primary challenge. Background shift refers to the semantic ambiguity of labels. To be specific, pixels labeled as background in the current task may in fact belong to other classes in previous or future tasks.
We implemented two segmentation-specific methods, ClassCL-PLOP and ClassCL-MiB, for comparison. Additionally, in order to reduce the influence of background shift, we introduced unbiased cross entropy loss in ClassCL-MiB for regularization-based and replay-based methods.

As Table \ref{tab:class-cl} shows, regularization-based methods failed to address background shift, as reflected by mean A-Dice scores close to the Non-CL baseline (average A-Dice 0.492 for regularization methods vs. 0.471 for Non-CL) and their WCD values remaining low (average WCD 0.562 vs. 0.572 for Non-CL). Replay-based methods achieved better results than regularization, although certain methods such as Repl-GSS and Repl-GPM exhibited limited general performance, with A-Dice scores of 0.456 and 0.575, respectively. This can be attributed to the lack of mechanisms explicitly designed to mitigate background shift in these strategies.
In contrast, parameter-isolation approaches delivered strong general performance, as their task-specific networks naturally address background shift by processing each task with independent parameters.

\textbf{Segmentation-specific methods achieved superior whole class segmentation performance with minor forgetting.}
In the Class-CL setting, ClassCL-PLOP and ClassCL-MiB delivered strong general performance, with A-Dice scores of 0.708 and 0.751, respectively. Both maintained low forgetting, with BWTR values of $–0.104$ for ClassCL-PLOP and $–0.039$ for ClassCL-MiB. Both methods delivered the top WCD scores, $0.707$ and $0.697$, indicating competitive whole-heart segmentation after the final task. This can be attributed to their balanced loss functions and pseudo-labeling strategies, which effectively mitigate background shift.


\subsection{Organ-CL Study} \label{sec:3.3}

\begin{table*}[!t]
\centering
\caption{Results of Organ-CL. An asterisk in the DRR column indicates that the method replays raw samples from previous tasks.}
\label{tab:Organ-CL}
\small
\renewcommand{\arraystretch}{1.2}
\begin{tabular}{@{}lccccc@{}}
\toprule
Method & A-Dice $\uparrow$ & BWTR $\uparrow$ & RMA $\uparrow$ & MPE $\downarrow$ & DRR $\downarrow$ \\
\midrule
Regu-EWC  & $.601\pm .008$ & $-0.371\pm .014$ & $1.004\pm .012$ & $0$ & $0$ \\
\hdashline
Regu-SI   & $.621\pm .024$ & $-0.343\pm .025$ & $1.003\pm .009$ & $0$ & $0$ \\
\hdashline
Regu-LwF  & $.594\pm .018$ & $-0.308\pm .044$ & $.916\pm .004$  & $0$ & $0$ \\
\midrule
Repl-ER   & $.729\pm .009$ & $-0.178\pm .021$ & $.996\pm .011$  & $0$ & $.012^{*}$ \\
\hdashline
Repl-GSS  & $.717\pm .005$ & $-0.276\pm .055$ & $.995\pm .009$  & $0$ & $.023^{*}$ \\
\hdashline
Repl-GEM  & $.599\pm .007$ & $-0.182\pm .068$ & $.806\pm .059$  & $0$ & $.023^{*}$ \\
\hdashline
Repl-GPM  & $.605\pm .009$ & $-0.329\pm .034$ & $.960\pm .027$  & $0$ & $.068$ \\
\hdashline
Repl-AGEM & $.635\pm .013$ & $-0.193\pm .069$ & $.873\pm .052$  & $0$ & $.023^{*}$ \\
\hdashline
Repl-DER  & $.717\pm .005$ & $-0.157\pm .022$ & $.953\pm .013$  & $0$ & $.012^{*}$ \\
\hdashline
Repl-FDR  & $.631\pm .008$ & $-0.217\pm .008$ & $.885\pm .013$  & $0$ & $.023^{*}$ \\
\midrule
ParIso-PNN & $.839\pm .004$ & $0$ & $1.001\pm .006$ & $1.074$ & $0$ \\
\hdashline
ParIso-DAN & $.824\pm .005$ & $0$ & $.972\pm .016$  & $.166$ & $0$ \\
\midrule
Non-CL    & $.603\pm .008$ & $-0.369\pm .005$ & $1.006\pm .008$ & $0\pm0$ & $0$ \\
\midrule
JointTrain & $.881\pm .010$ & / & / & $0$ & $0$ \\
\bottomrule
\end{tabular}
\end{table*}

Similar to the Domain-CL setting, weight regularization failed to mitigate forgetting in Organ-CL. Regu-EWC, Regu-SI and Regu-LwF showed BWTR values near the Non-CL baseline, i.e., $-0.371$, $-0.343$ and $-0.308$, with A-Dice of $0.601$, $0.621$ and $0.594$. 
Replay-based methods such as Repl-ER and Repl-DER achieved better stability, with BWTR values of $–0.178$ and $–0.157$, while maintaining high plasticity with RMA scores of $0.996$ and $0.953$, respectively.
Parameter-isolation methods completely eliminated forgetting and achieved the highest A-Dice scores, with ParIso-PNN at $0.839$ and ParIso-DAN at $0.824$. In addition, the parameter-isolation strategy showed high learning plasticity, with ParIso-PNN and ParIso-DAN reaching RMA scores of $1.001$ and $0.972$, respectively. Similar to the Domain-CL setting, ParIso-DAN exhibited higher parameter efficiency with 0.166 MPE score compared to ParIso-PNN.

\subsection{Study of Task-ordering Robustness} \label{sec:3.4}

\begin{figure*}[!t]
    \centering

    \begin{subfigure}{0.78\textwidth}
        \centering
        \includegraphics[width=\linewidth]{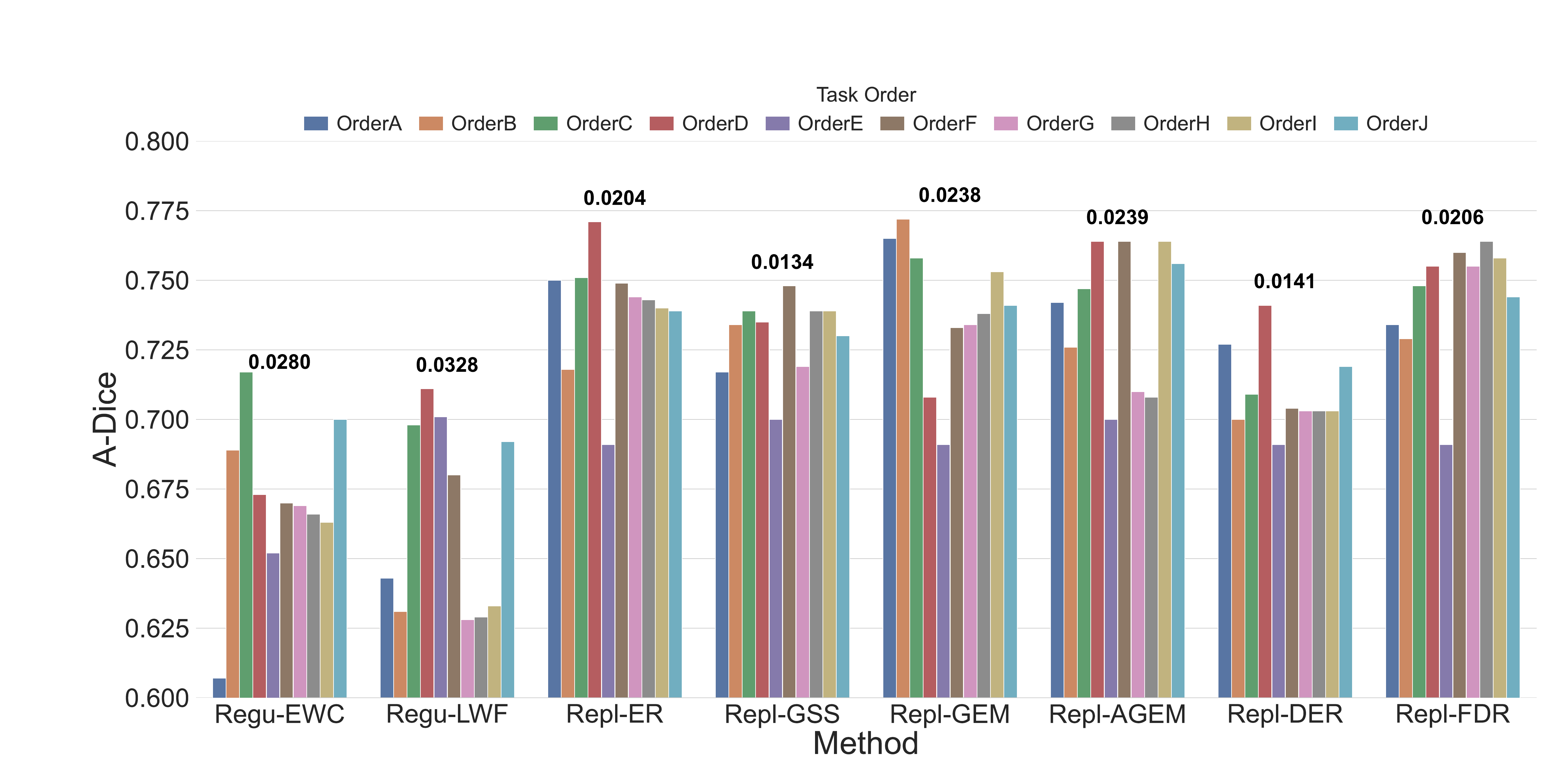}
        \caption{A-Dice}
        \label{fig:acc_robustness}
    \end{subfigure}

    \vspace{0.1em}

    \begin{subfigure}{0.78\textwidth}
        \centering
        \includegraphics[width=\linewidth]{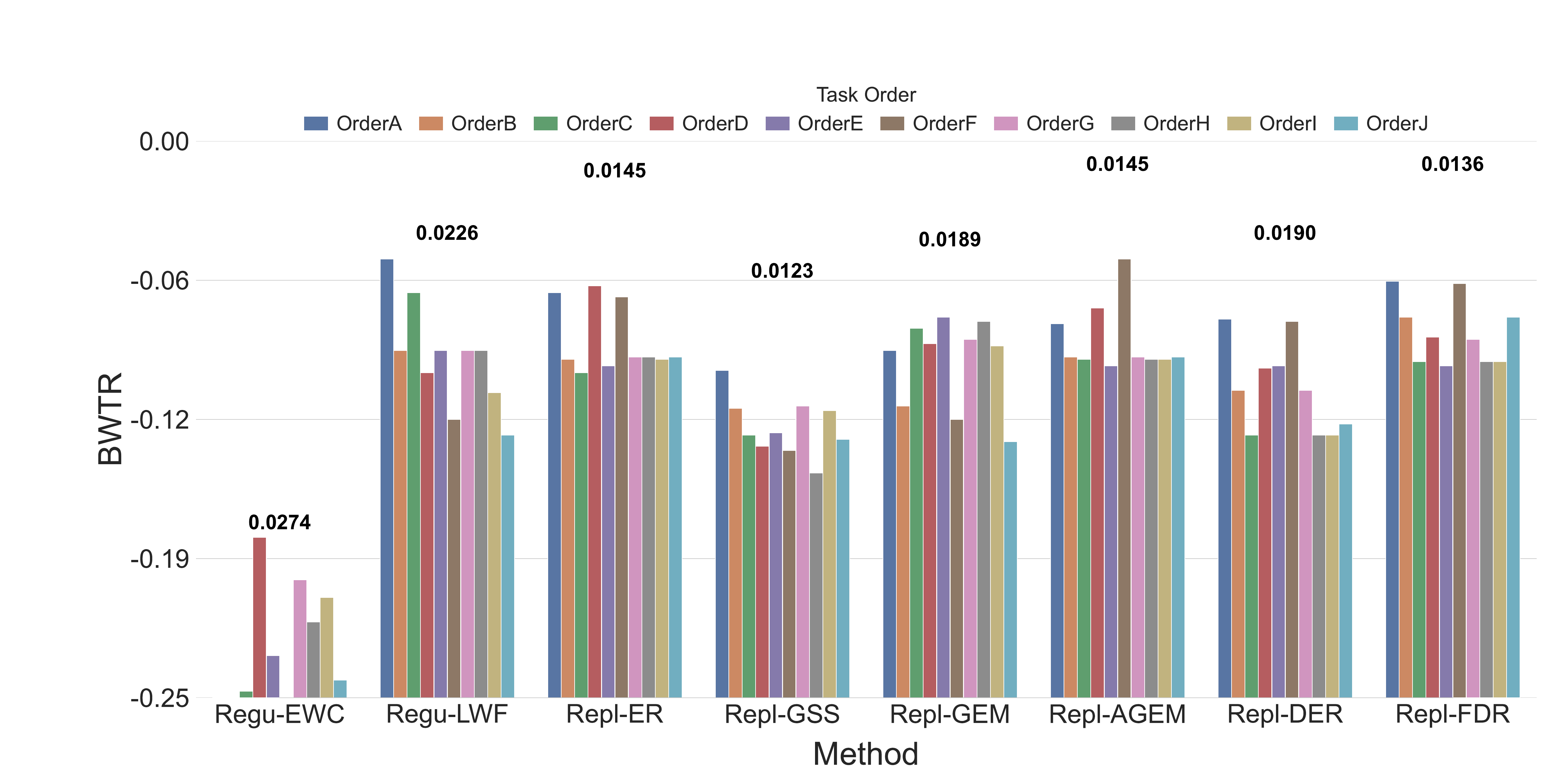}
        \caption{BWTR}
        \label{fig:BWTR_robustness}
    \end{subfigure}

    \caption{Task-order robustness performance of CL methods. Subfigures show (a) A-Dice and (b) BWTR results with standard deviations across ten different task orders.}
    \label{fig:task_order_robustness}
\end{figure*}

We present a task-ordering robustness study on the Domain-CL setting. 
Task-ordering in CL may affect both individual task and overall performance, due to factors like data similarity across different tasks and catastrophic forgetting. 
Therefore, it is important to investigate the influence of task order on chosen methods in continual medical image segmentation.
Specifically, we selected ten random permutations of task sequences, denoted as OrderA to OrderJ. We then report the general performance and stability of both regularization-based and replay-based methods. As Fig.~\ref{fig:task_order_robustness} shows, both A-Dice and BWTR present variations across ten different task orders, confirming that task-ordering affects CL performance. 

\textbf{Replay-based methods are robust to task-order changes and buffer data diversity improves robustness.}
Regularization-based methods, Regu-EWC and Regu-LwF, showed the largest performance variations, while replay-based methods exhibit lower performance variations. Moreover, among the evaluated methods, Repl-GSS consistently shows the highest robustness with the lowest standard deviations in both A-Dice (0.0134) and BWTR (0.0123). This can be attributed to the sampling strategy in Repl-GSS, which explicitly considers the diversity of stored buffer samples during replay.

\begin{figure*}[!t]
\centering

\subfloat[Effect of buffer size for replay-based methods.]{
  \includegraphics[width=0.82\textwidth]{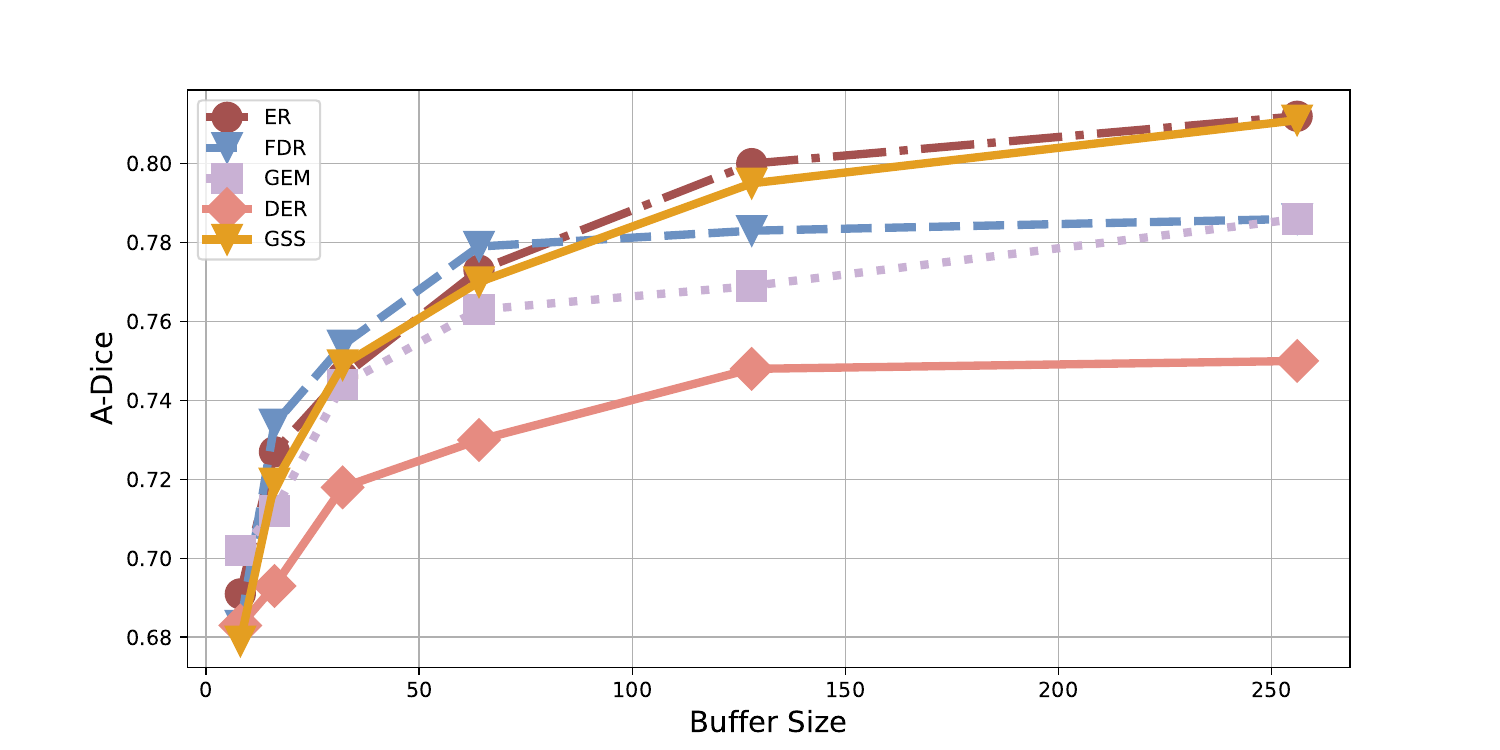}
  \label{fig:memory_a}
}

\vspace{1em}

\subfloat[Task-wise Dice matrix of SAM and SAM with LoRA fine-tuning.]{
  \includegraphics[width=0.83\textwidth]{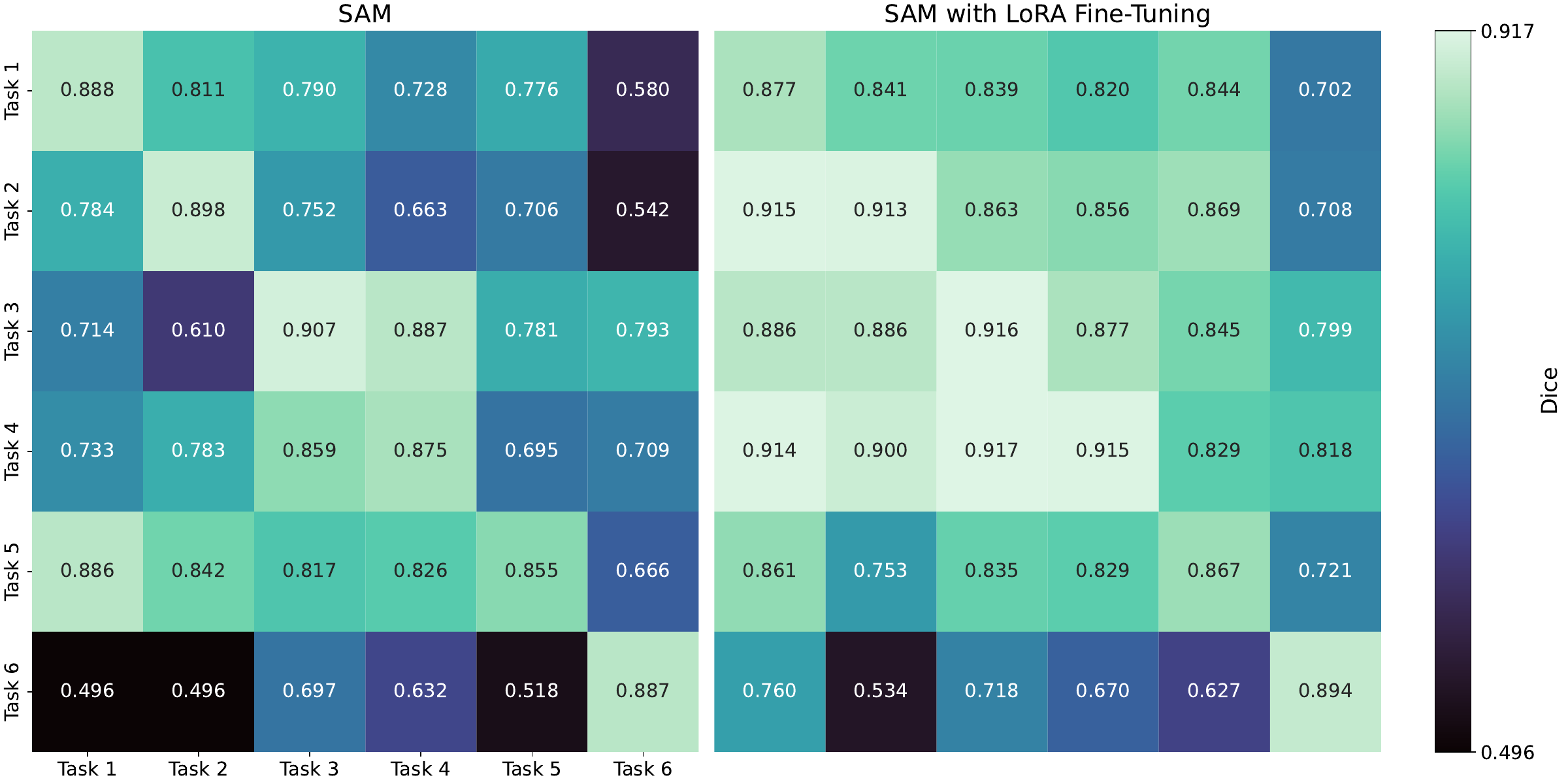}
  \label{fig:memory_b}
}

\caption{Additional analyses of replay-based methods and foundation-model behavior.
(a) Effect of buffer size on A-Dice for replay-based methods.
(b) Task-wise Dice matrices of vanilla sequential SAM fine-tuning and sequential SAM with LoRA under the Domain-CL setting. Rows denote the model after completing training on each task, and columns denote evaluation tasks. In vanilla sequential SAM fine-tuning, both encoder and decoder are updated. In sequential SAM with LoRA, the encoder is frozen, while LoRA modules and the decoder are updated.}
\label{fig:memory}
\end{figure*}

\subsection{Study of Buffer Size for Replay-based Methods} \label{sec:3.5}
The buffer size in replay-based methods is crucial for the effectiveness in retaining old knowledge and learning new tasks. Fig.~\ref{fig:memory} illustrates the impact of buffer size for representative replay-based methods.

\textbf{Increasing buffer size improves general performance, but the gains saturate as the buffer grows.}
As Fig.~\ref{fig:memory} shows, a positive correlation between buffer size and A-Dice is evident across all methods, indicating that increased buffer allocation generally enhances model performance. However, as the buffer size continues to grow, most methods exhibit a saturation effect, where performance improvements decrease. This effect arises because additional samples often carry redundant information, biasing the model toward previous tasks, increasing interference with the current task and compromising its learning. In addition, large buffer sizes introduce substantial computational and storage overhead, posing challenges in resource-limited settings. These findings highlight the need for more memory-efficient replay mechanisms that sustain performance without relying on large buffer sizes.

\begin{figure*}[t]
\centering
\includegraphics[width=0.75\textwidth]{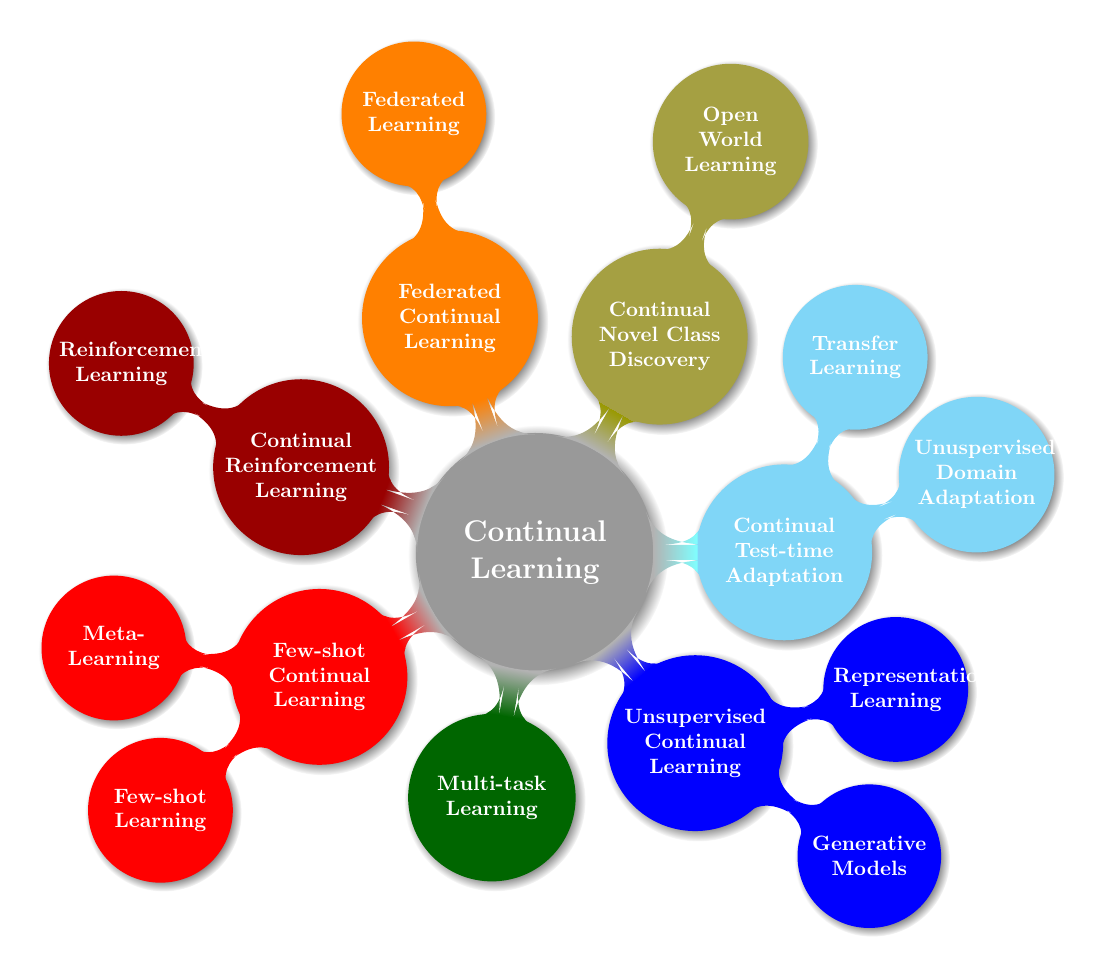}
\caption{CL and its connections with related machine learning paradigms.}
\label{fig:figure_machine_learning}
\end{figure*}

\section{Discussion and Conclusion}

Our benchmark highlights several central findings for continual medical image segmentation. First, performance remains constrained by multiple competing objectives rather than forgetting alone. Replay-based methods provide the most balanced trade-off between stability and plasticity, but they rely on replayed information that may be undesirable in privacy-sensitive settings. Parameter-isolation methods are most effective at eliminating forgetting, but this advantage comes with continual parameter growth. Moreover, forward generalizability remains limited across current strategies, indicating that reducing forgetting alone is not sufficient for real clinical deployment. These findings suggest several important directions for future work, particularly in addressing privacy constraints, limited annotation, and adaptation under evolving data distributions.

\subsection{Related Paradigms and Extensions of Continual Learning}

The integration of CL with related learning paradigms could lead to new approaches for addressing realistic challenges in medical image segmentation. Despite their significance, these directions have received limited attention in current research. Fig. ~\ref{fig:figure_machine_learning} illustrates the relationships between CL and related machine learning fields and the intersections. Here we discuss the following directions:



\textbf{Federated Continual Learning.}
Federated learning enables clinical centers to collaboratively train models without sharing sensitive medical data. However, the evolving nature of real-world clinical environments in local centers challenges the efficacy of traditional static federated learning\cite{discussion_fedcl_if, discussion_fedcl_icml2021, discussion_fedcl_pami}. 
Federated continual learning (FCL) emerges as an approach for learning from spatially distributed and temporally evolving medical data, enabling practical deployment of federated learning in real-world scenarios.
It represents a natural extension of CL by integrating learning across space and time within a unified framework. It has the potential to serve as the foundational strategy for developing public AI-based healthcare systems \cite{back_federated_medicine5, back_federated_medicine, back_federated_medicine2, back_federated_medicine3,back_federated_medicine4}.
In addition, substantial inconsistencies in disease progression and computational infrastructure exist among medical centers. 
As a result, FCL methods should facilitate asynchronous learning to address such real-world challenge\cite{discussion_fedcl_asy}.
 
\textbf{Continual Test-time Adaptation.} Test-time adaptation allows models to adjust to new data distributions using only unlabeled test samples, thereby avoiding the need for segmentation labeling or access to source training data. \cite{discussion_tta_tent, discussion_tta_tentseg}. This direction is especially relevant because our benchmark shows that current CL methods remain weak in forward transfer, while real clinical deployment often requires adaptation before dense annotations become available. Combined with CL properties, continual test-time adaptation can quickly adapt models to dynamic unlabeled test data, while preserving segmentation performance on source training data and previous test data.
Since collecting large-scale pixel-wise labeled data for medical image segmentation is laborious and time-consuming \cite{discussion_uda_review, discussion_uda_review2}, this approach has the potential to support efficient deployment of trained segmentation models. This approach holds potential for highly efficient deployment of trained segmentation models, enabling precision medicine and supporting real-time clinical workflows.


\subsection{Discussion of Foundation Models in CL}

The limited forward generalizability observed in our benchmark raises an important question: whether stronger pretrained representations can improve continual segmentation under evolving clinical conditions. Foundation models are particularly relevant in this context because they provide broad feature priors learned from large-scale data. However, their use in continual medical image segmentation is not straightforward. A foundation model must not only adapt to new tasks, but also preserve previous knowledge, control parameter growth, and maintain generalizability to future domains.

Foundation models such as the Segment Anything Model (SAM)~\cite{foundation_sam} have shown strong potential for medical image segmentation. Their large-scale pretraining enables broad feature transfer across heterogeneous datasets and segmentation targets~\cite{back_segmentation_sam, foundation_nature, foundation_nature2, foundation_nature3, foundation_nature4, foundation_nature5}. Nevertheless, foundation-model priors alone do not remove the core challenges of continual learning. To provide a preliminary analysis, we evaluate SAM in the Domain-CL setting under two sequential adaptation strategies. In the first strategy, we start from SAM and sequentially fine-tune it on Task 1, Task 2, ..., Task 6, with both the encoder and decoder updated. In the second strategy, we start from the same SAM initialization and perform sequential LoRA fine-tuning, where the encoder is frozen and only the LoRA modules and decoder are updated.

Fig.~\ref{fig:memory}(b) reports the resulting task-wise Dice matrices. In each matrix, the $i$-th row denotes the model after completing training on Task $i$, and the columns denote evaluation tasks across all six Domain-CL domains. Thus, the diagonal entries reflect adaptation to the current task, the lower-triangular entries reflect retention of previous tasks, and the upper-triangular entries reflect generalization to future unseen domains. This design directly evaluates the continual behavior of foundation-model adaptation rather than isolated task-specific fine-tuning.

The vanilla sequential SAM fine-tuning matrix shows strong performance on several diagonal entries, indicating that full fine-tuning can adapt SAM to individual domains. However, its off-diagonal entries remain uneven, and performance on previous domains drops substantially after later adaptation. For example, after training on Task 6, the average Dice over previous tasks is markedly reduced. This suggests that full fine-tuning can overwrite useful pretrained representations and does not reliably preserve cross-domain generalizability in a continual setting.

Sequential SAM with LoRA shows a more favorable pattern. Compared with vanilla sequential fine-tuning, LoRA improves the overall task-wise Dice matrix and yields stronger off-diagonal performance. This improvement indicates better cross-task transfer and retention, rather than only better fitting to the current task. In this example, the average Dice increases from 0.747 to 0.823, and the average off-diagonal Dice increases from 0.719 to 0.808. These results suggest that parameter-efficient adaptation can better exploit foundation-model priors while reducing the risk of destructive full-model updates.

This observation supports the use of parameter-efficient fine-tuning methods~\cite{discussion_lora, discussion_peft_survey} for continual medical image segmentation. LoRA~\cite{discussion_lora} introduces trainable low-rank matrices into the Transformer architecture while keeping most pretrained weights fixed. In a continual setting, this design is attractive because it allows task adaptation with limited parameter updates, while preserving a stable pretrained representation. However, LoRA does not fully solve continual learning. Since the decoder and task-specific adaptation modules are still updated sequentially, forgetting and parameter accumulation may still occur as the number of tasks increases.

These findings point to several directions for foundation-model-based continual segmentation. First, task-specific or domain-specific LoRA modules can be expanded over time to support new clinical tasks while preserving the pretrained encoder. Second, adapter merging, adapter selection, or adapter pruning could be used to control parameter growth when the number of tasks increases. This is important because simply adding one adapter for each new task may eventually reproduce the scalability problem of parameter-isolation methods. Third, foundation models can be combined with conventional CL strategies. For instance, regularization can constrain changes in adapter or decoder parameters, replay-free subspace methods can protect important update directions, and lightweight task-specific heads can decouple anatomical targets while sharing a common encoder.

In summary, foundation models provide useful pretrained representations for continual medical image segmentation, but they do not by themselves eliminate forgetting or guarantee forward generalizability. The Domain-CL analysis in Fig.~\ref{fig:memory}(b) suggests that parameter-efficient sequential adaptation, such as LoRA fine-tuning, is a promising starting point. Future work should further investigate how adapters can be selected, merged, regularized, or shared across evolving clinical tasks while maintaining both parameter efficiency and cross-domain robustness.




\subsection{Conclusion and Future Perspectives}
In this paper, we construct a benchmark for CL in medical image segmentation, comprising three CL scenarios with corresponding datasets. We further introduce new metrics for comprehensive CL performance evaluation. 
We also conduct a benchmark study of representative CL methods and provide empirical insights into their performance.
We further investigate key factors affecting CL performance, including task order and memory size. 
Finally, we discuss the potential of foundation models in continual medical image segmentation. 


Here, we discuss the following perspectives on future research trends in CL for medical image segmentation:

\textbf{Multi-modality.} Multi-modal large language models (MMLLMs) have the potential to achieve knowledge-enhanced medical image segmentation by integrating and processing textual information. These models combine textual and image data, allowing them to capture correlations between medical images and accompanying clinical knowledge and to support segmentation across modalities and anatomies.


\textbf{Interpretability.} Interpretability enables researchers and clinicians to understand the conditions under which AI models perform effectively and identify potential failure cases, thereby enhancing the safety, reliability, and transparency of AI systems \cite{discussion_inter_cls, discussion_inter_cls2, discussion_inter_seg, discussion_inter_seg2, discussion_inter_seg3, discussion_inter_seg4}. A recent study \cite{discussion_inter_covid} reveals that DL systems detecting COVID-19 from radiographs often rely on confounding factors, demonstrating that explainable AI should be a prerequisite for clinical deployment. In the context of CL for medical image segmentation, improving the interpretability of CL methods is essential for future clinical applications. 

\textbf{Bio-inspired Modeling.} Biological learning systems are naturally endowed with the ability to learn continually. Its underlying mechanisms offer valuable insights and AI practitioners can draw inspiration from it in developing CL approaches \cite{discussion_computational_models, back_biological_brain_active_forgetting}. We claim that continual learning (CL) strategies should extend their focus beyond simply addressing catastrophic forgetting, aiming instead for systems capable of learning, memorizing, retrieving, and transferring knowledge akin to the human brain. 

\textbf{Active Forgetting.} Recent studies view forgetting as an essential aspect of brain function. Active forgetting could release the learning ability for new tasks \cite{back_biological_brain_active_forgetting}. Active forgetting plays a crucial role in continual learning for medical image segmentation by enabling models to selectively discard outdated or irrelevant information, thus enhancing adaptability and efficiency. However, implementing active forgetting presents challenges, including determining what knowledge to forget and maintaining a balance between retaining valuable information and discarding unnecessary information. These challenges are particularly significant in medical settings, where decisions on data retention and forgetting must be carefully managed to avoid adverse impacts on patient care.

\section*{Acknowledgments}
This work was funded by 
the Science and Technology Commission of Shanghai Municipality (25TS1412100),
the Noncommunicable Chronic Disease-National Science and Technology Major Project (2026ZD0555800/ 2026ZD0555802), 
and the National Natural Science Foundation of China (62372115).


\begin{IEEEbiography}[{\includegraphics[width=1in,height=1.25in,clip,keepaspectratio]{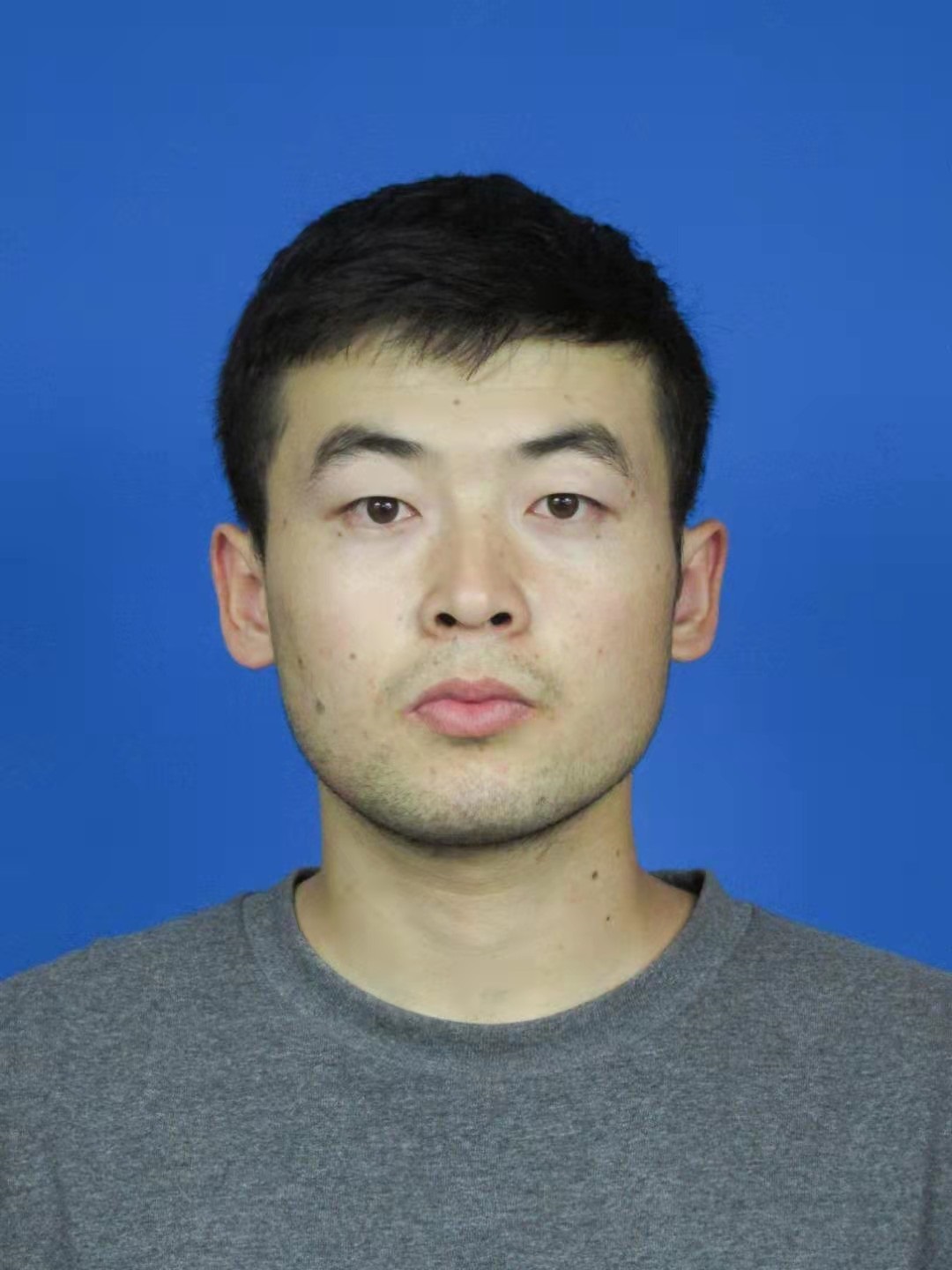}}]{Bomin Wang}
is a Ph.D. candidate at the School of Data Science, Fudan University, under the supervision of Prof. Xiahai Zhuang. He received his Bachelor's degree from Lanzhou University and Master's degree from Shandong University. His current research interests focus on continual learning and medical image analysis.
\end{IEEEbiography}

\begin{IEEEbiography}[{\includegraphics[width=1in,height=1.25in,clip,keepaspectratio]{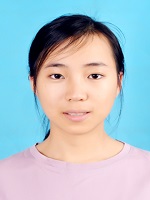}}]{Hangqi Zhou}
is doing a Ph.D. at the School of Data Science, Fudan University, Shanghai, China, with Prof. Xiahai Zhuang. She received her Bachelor's degree from Fudan University. Her current research interests focus on interpretable deep learning, continual learning and medical image analysis.
\end{IEEEbiography}

\begin{IEEEbiography}[{\includegraphics[width=1in,height=1.25in,clip,keepaspectratio]{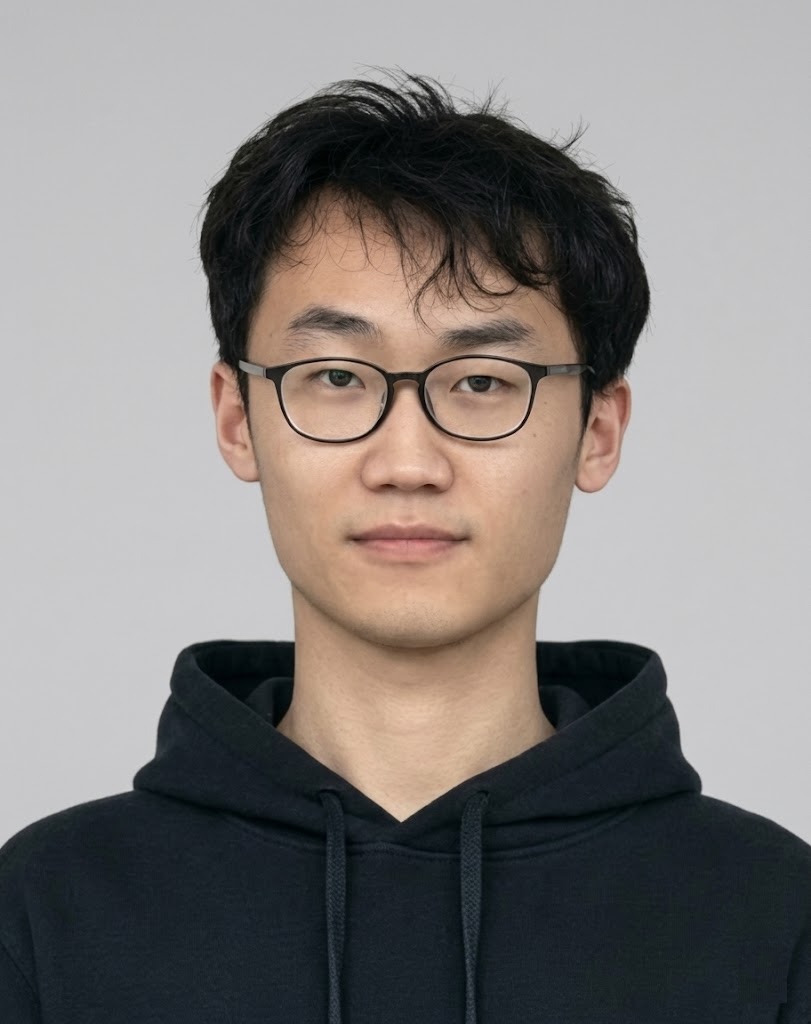}}]{Yibo Gao}
is a Ph.D. candidate at the School of Data Science, Fudan University, under the supervision of Prof. Xiahai Zhuang. He received his B.E. degree from the University of Electronic Science and Technology of China. His research interests include interpretable AI and medical image analysis. His work won the Elsevier-MedIA 1st Prize and Medical image Analysis MICCAI Best Paper Award in 2023. He also received the MICCAI Young Scientist Award in 2025.
\end{IEEEbiography}

\begin{IEEEbiography}[{\includegraphics[width=1in,height=1.25in,clip,keepaspectratio]{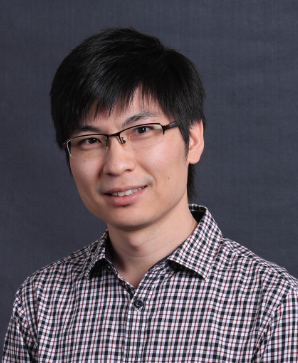}}]{Xiahai Zhuang}
is a professor at School of Data Science, Fudan University. He graduated from the Department of Computer Science, Tianjin University, received a Master's degree from Shanghai Jiao Tong University, and a Doctorate degree from University College London. His research interests include interpretable Al, medical image analysis and computer vision. His work won the Elsevier-MedIA 1st Prize and Medical image Analysis MICCAI Best Paper Award 2023.
\end{IEEEbiography}


\begin{thebibliography}{00}



\bibitem{back_biological_brain_active_forgetting} Wang L, Zhang X, Li Q, et al. Incorporating neuro-inspired adaptability for continual learning in artificial intelligence[J]. Nature Machine Intelligence, 2023, 5(12): 1356-1368.



\bibitem{discussion_computational_models} Gluck M A, Granger R. Computational models of the neural bases of learning and memory[J]. Annual review of neuroscience, 1993, 16(1): 667-706.


\bibitem{back_AI_for_medicine} Rajpurkar P, Chen E, Banerjee O, et al. AI in health and medicine[J]. Nature medicine, 2022, 28(1): 31-38.

\bibitem{back_AI_for_medicine_retina} Gulshan V, Peng L, Coram M, et al. Development and validation of a deep learning algorithm for detection of diabetic retinopathy in retinal fundus photographs[J]. Jama, 2016, 316(22): 2402-2410.


\bibitem{back_survey_defying} De Lange M, Aljundi R, Masana M, et al. A continual learning survey: Defying forgetting in classification tasks[J]. IEEE transactions on pattern analysis and machine intelligence, 2021, 44(7): 3366-3385.

\bibitem{back_survey_liyuan} Wang L, Zhang X, Su H, et al. A comprehensive survey of continual learning: Theory, method and application[J]. IEEE Transactions on Pattern Analysis and Machine Intelligence, 2024.

\bibitem{back_survey_transai} Wickramasinghe B, Saha G, Roy K. Continual learning: A review of techniques, challenges, and future directions[J]. IEEE Transactions on Artificial Intelligence, 2023, 5(6): 2526-2546.

\bibitem{back_clinical_cl} Lee C S, Lee A Y. Clinical applications of continual learning machine learning[J]. The Lancet Digital Health, 2020, 2(6): e279-e281.

\bibitem{back_distribution_shift_nips} Li H, Wang Y F, Wan R, et al. Domain generalization for medical imaging classification with linear-dependency regularization[J]. Advances in neural information processing systems, 2020, 33: 3118-3129.


\bibitem{replay_HIPAA} Act A. Health insurance portability and accountability act of 1996[J]. Public law, 1996, 104: 191.

\bibitem{replay_GDPR} Regulation P. General data protection regulation[J]. Intouch, 2018, 25: 1-5.

\bibitem{back_distribution_shift_biostat} Subbaswamy A, Saria S. From development to deployment: dataset shift, causality, and shift-stable models in health AI[J]. Biostatistics, 2020, 21(2): 345-352.






\bibitem{back_AI_for_COVID} Jin C, Chen W, Cao Y, et al. Development and evaluation of an artificial intelligence system for COVID-19 diagnosis[J]. Nature Communications, 2020, 11(1): 5088.



\bibitem{back_segmentation_sam} Ma J, He Y, Li F, et al. Segment anything in medical images[J]. Nature Communications, 2024, 15(1): 654.






\bibitem{method_replay_dgr} Shin H, Lee J K, Kim J, et al. Continual learning with deep generative replay[J]. Advances in neural information processing systems, 2017, 30.







\bibitem{method_regu_ewc} Kirkpatrick J, Pascanu R, Rabinowitz N, et al. Overcoming catastrophic forgetting in neural networks[J]. Proceedings of the national academy of sciences, 2017, 114(13): 3521-3526.

\bibitem{method_regu_si} Zenke F, Poole B, Ganguli S. Continual learning through synaptic intelligence[C]//International conference on machine learning. PMLR, 2017: 3987-3995.


\bibitem{method_regu_lwf} Li Z, Hoiem D. Learning without forgetting[J]. IEEE Transactions on Pattern Analysis and Machine Intelligence, 2017, 40(12): 2935-2947.

\bibitem{dataset_lascars2} Li L, Zimmer V A, Schnabel J A, et al. Medical image analysis on left atrial LGE MRI for atrial fibrillation studies: A review[J]. Medical Image Analysis, 2022, 77: 102360.

\bibitem{method_replay_fdr} Benjamin A, Rolnick D, Kording K. Measuring and regularizing networks in function space[C]//International Conference on Learning Representations.



\bibitem{method_dyn_wsn} Kang H, Mina R J L, Madjid S R H, et al. Forget-free continual learning with winning subnetworks[C]//International Conference on Machine Learning. PMLR, 2022: 10734-10750.

\bibitem{method_dyn_pnn} Rusu A A, Rabinowitz N C, Desjardins G, et al. Progressive neural networks[J]. arXiv preprint arXiv:1606.04671, 2016.

\bibitem{method_dyn_dan} Rosenfeld A, Tsotsos J K. Incremental learning through deep adaptation[J]. IEEE transactions on pattern analysis and machine intelligence, 2020, 42(3): 651-663.

\bibitem{method_replay_er} Rolnick D, Ahuja A, Schwarz J, et al. Experience replay for continual learning[J]. Advances in neural information processing systems, 2019, 32.


\bibitem{method_replay_gss} Tiwari R, Killamsetty K, Iyer R, et al. Gcr: Gradient coreset based replay buffer selection for continual learning[C]//Proceedings of the IEEE/CVF Conference on Computer Vision and Pattern Recognition. 2022: 99-108.


\bibitem{dataset_lascars} Li L, Zimmer V A, Schnabel J A, et al. AtrialJSQnet: a new framework for joint segmentation and quantification of left atrium and scars incorporating spatial and shape information[J]. Medical Image Analysis, 2022, 76: 102303.

\bibitem{method_replay_mer} Riemer M, Cases I, Ajemian R, et al. Learning to Learn without Forgetting by Maximizing Transfer and Minimizing Interference[C]//International Conference on Learning Representations.

\bibitem{method_replay_gem} Lopez-Paz D, Ranzato M A. Gradient episodic memory for continual learning[J]. Advances in neural information processing systems, 2017, 30.

\bibitem{method_replay_agem} Chaudhry A, Ranzato M A, Rohrbach M, et al. Efficient Lifelong Learning with A-GEM[C]//International Conference on Learning Representations. 2018.


\bibitem{method_regu_gpm} Saha G, Garg I, Roy K. Gradient Projection Memory for Continual Learning[C]//International Conference on Learning Representations.





\bibitem{method_replay_der} Buzzega P, Boschini M, Porrello A, et al. Dark experience for general continual learning: a strong, simple baseline[J]. Advances in neural information processing systems, 2020, 33: 15920-15930.

\bibitem{method_replay_xder} Boschini M, Bonicelli L, Buzzega P, et al. Class-incremental continual learning into the extended der-verse[J]. IEEE Transactions on Pattern Analysis and Machine Intelligence, 2022, 45(5): 5497-5512.

\bibitem{method_seg_mib} Cermelli F, Mancini M, Bulo S R, et al. Modeling the background for incremental learning in semantic segmentation[C]//Proceedings of the IEEE/CVF Conference on Computer Vision and Pattern Recognition. 2020: 9233-9242.

\bibitem{method_seg_plop} Douillard A, Chen Y, Dapogny A, et al. Plop: Learning without forgetting for continual semantic segmentation[C]//Proceedings of the IEEE/CVF conference on computer vision and pattern recognition. 2021: 4040-4050.



\bibitem{dataset_domaincl} Liu Q, Dou Q, Heng P A. Shape-aware meta-learning for generalizing prostate MRI segmentation to unseen domains[C]//Medical Image Computing and Computer Assisted Intervention–MICCAI 2020: 23rd International Conference, Lima, Peru, October 4–8, 2020, Proceedings, Part II 23. Springer International Publishing, 2020: 475-485.


\bibitem{care25ASA1} Zhuang X, Shen J. Multi-scale patch and multi-modality atlases for whole heart segmentation of MRI[J]. Medical image analysis, 2016, 31: 77-87.

\bibitem{care25ASA2} Zhuang X. Multivariate mixture model for myocardial segmentation combining multi-source images[J]. IEEE transactions on pattern analysis and machine intelligence, 2018, 41(12): 2933-2946.



\bibitem{dataset_taskcl} Bilic P, Christ P, Li H B, et al. The liver tumor segmentation benchmark (lits)[J]. Medical Image Analysis, 2023, 84: 102680.

\bibitem{dataset_taskcl2} Pati S, Baid U, Zenk M, et al. The federated tumor segmentation (fets) challenge[J]. arXiv preprint arXiv:2105.05874, 2021.

\bibitem{dataset_prostate} Lemaître G, Martí R, Freixenet J, et al. Computer-aided detection and diagnosis for prostate cancer based on mono and multi-parametric MRI: a review[J]. Computers in biology and medicine, 2015, 60: 8-31.

\bibitem{dataset_prostate2} Bloch, N., Madabhushi, A., Huisman, H., Freymann, J., et al.: NCI-ISBI 2013 Challenge: Automated Segmentation of Prostate Structures, 2014.

\bibitem{dataset_prostate3} Litjens, G., Toth, R., Ven, W., Hoeks, C., et al.: Evaluation of prostate segmentation algorithms for mri: The promise12 challenge[J]. Medical Image Analysis, 2014, 18:359-373.

\bibitem{back_federated_medicine} Dayan I, Roth H R, Zhong A, et al. Federated learning for predicting clinical outcomes in patients with COVID-19[J]. Nature medicine, 2021, 27(10): 1735-1743.

\bibitem{back_federated_medicine2} Rieke N, Hancox J, Li W, et al. The future of digital health with federated learning[J]. NPJ digital medicine, 2020, 3(1): 1-7.

\bibitem{back_federated_medicine3} Froelicher D, Troncoso-Pastoriza J R, Raisaro J L, et al. Truly privacy-preserving federated analytics for precision medicine with multiparty homomorphic encryption[J]. Nature Communications, 2021, 12(1): 5910.

\bibitem{back_federated_medicine4} Ogier du Terrail J, Leopold A, Joly C, et al. Federated learning for predicting histological response to neoadjuvant chemotherapy in triple-negative breast cancer[J]. Nature Medicine, 2023, 29(1): 135-146.

\bibitem{back_federated_medicine5} Sadilek A, Liu L, Nguyen D, et al. Privacy-first health research with federated learning[J]. NPJ digital medicine, 2021, 4(1): 132.


\bibitem{discussion_fedcl_icml2021} Yoon J, Jeong W, Lee G, et al. Federated continual learning with weighted inter-client transfer[C]//International Conference on Machine Learning. PMLR, 2021: 12073-12086.

\bibitem{discussion_fedcl_if} Criado M F, Casado F E, Iglesias R, et al. Non-iid data and continual learning processes in federated learning: A long road ahead[J]. Information Fusion, 2022, 88: 263-280.

\bibitem{discussion_fedcl_pami} Dong J, Li H, Cong Y, et al. No one left behind: Real-world federated class-incremental learning[J]. IEEE Transactions on Pattern Analysis and Machine Intelligence, 2023.

\bibitem{discussion_fedcl_asy} Shenaj D, Toldo M, Rigon A, et al. Asynchronous federated continual learning[C]//Proceedings of the IEEE/CVF Conference on Computer Vision and Pattern Recognition. 2023: 5055-5063.

\bibitem{discussion_tta_tent} Wang D, Shelhamer E, Liu S, et al. Tent: Fully Test-Time Adaptation by Entropy Minimization[C]//International Conference on Learning Representations.

\bibitem{discussion_tta_tentseg} Hu M, Song T, Gu Y, et al. Fully test-time adaptation for image segmentation[C]//Medical Image Computing and Computer Assisted Intervention–MICCAI 2021: 24th International Conference, Strasbourg, France, September 27–October 1, 2021, Proceedings, Part III 24. Springer International Publishing, 2021: 251-260.




\bibitem{discussion_uda_review} Toldo M, Maracani A, Michieli U, et al. Unsupervised domain adaptation in semantic segmentation: a review[J]. Technologies, 2020, 8(2): 35.



\bibitem{discussion_lora} Hu E J, Shen Y, Wallis P, et al. Lora: Low-rank adaptation of large language models[J]. ICLR, 2022, 1(2): 3.

\bibitem{discussion_uda_review2} Csurka G, Volpi R, Chidlovskii B. Unsupervised domain adaptation for semantic image segmentation: a comprehensive survey[J]. arXiv preprint arXiv:2112.03241, 2021.








\bibitem{foundation_nature} Moor M, Banerjee O, Abad Z S H, et al. Foundation models for generalist medical artificial intelligence[J]. Nature, 2023, 616(7956): 259-265.

\bibitem{foundation_nature2} Zhou Y, Chia M A, Wagner S K, et al. A foundation model for generalizable disease detection from retinal images[J]. Nature, 2023, 622(7981): 156-163.

\bibitem{foundation_nature3} Fei N, Lu Z, Gao Y, et al. Towards artificial general intelligence via a multimodal foundation model[J]. Nature Communications, 2022, 13(1): 3094.

\bibitem{foundation_nature4} Cui H, Wang C, Maan H, et al. scGPT: toward building a foundation model for single-cell multi-omics using generative AI[J]. Nature Methods, 2024: 1-11.

\bibitem{foundation_nature5} Guo L L, Fries J, Steinberg E, et al. A multi-center study on the adaptability of a shared foundation model for electronic health records[J]. npj Digital Medicine, 2024, 7(1): 171.









\bibitem{discussion_peft_survey} Ding N, Qin Y, Yang G, et al. Parameter-efficient fine-tuning of large-scale pre-trained language models[J]. Nature Machine Intelligence, 2023, 5(3): 220-235.


















\bibitem{foundation_sam} Kirillov A, Mintun E, Ravi N, et al. Segment anything[C]//Proceedings of the IEEE/CVF International Conference on Computer Vision. 2023: 4015-4026.













\bibitem{discussion_inter_seg} Sun J, Darbehani F, Zaidi M, et al. Saunet: Shape attentive u-net for interpretable medical image segmentation[C]//Medical Image Computing and Computer Assisted Intervention–MICCAI 2020: 23rd International Conference, Lima, Peru, October 4–8, 2020, Proceedings, Part IV 23. Springer International Publishing, 2020: 797-806.

\bibitem{discussion_inter_seg2} Zhang Z, Xie Y, Xing F, et al. Mdnet: A semantically and visually interpretable medical image diagnosis network[C]//Proceedings of the IEEE conference on computer vision and pattern recognition. 2017: 6428-6436.

\bibitem{discussion_inter_seg3} Gao S, Zhou H, Gao Y, et al. BayeSeg: Bayesian modeling for medical image segmentation with interpretable generalizability[J]. Medical Image Analysis, 2023, 89: 102889.

\bibitem{discussion_inter_seg4} He S, Feng Y, Grant P E, et al. Segmentation ability map: Interpret deep features for medical image segmentation[J]. Medical image analysis, 2023, 84: 102726.

\bibitem{discussion_inter_cls} He S, Feng Y, Grant P E, et al. Chanda T, Hauser K, Hobelsberger S, et al. Dermatologist-like explainable AI enhances trust and confidence in diagnosing melanoma[J]. Nature Communications, 2024, 15(1): 524.

\bibitem{discussion_inter_cls2} Kim C, Gadgil S U, DeGrave A J, et al. Transparent medical image AI via an image–text foundation model grounded in medical literature[J]. Nature Medicine, 2024: 1-12.

\bibitem{discussion_inter_covid} DeGrave A J, Janizek J D, Lee S I. AI for radiographic COVID-19 detection selects shortcuts over signal[J]. Nature Machine Intelligence, 2021, 3(7): 610-619.

\bibitem{lifelong_unet} González C, Ranem A, Pinto dos Santos D, et al. Lifelong nnU-Net: a framework for standardized medical continual learning[J]. Scientific Reports, 2023, 13(1): 9381.

\bibitem{whatwrong} Gonzalez C, Lemke N, Ranem A, et al. What is wrong with continual learning in medical image segmentation?[C]//Proceedings of the International Workshop on Personalized Incremental Learning in Medicine. 2025: 25-34.

\bibitem{medcl_miccai22} Liu P, Wang X, Fan M, et al. Learning incrementally to segment multiple organs in a CT image[C]//International Conference on Medical Image Computing and Computer-Assisted Intervention. Cham: Springer Nature Switzerland, 2022: 714-724.

\bibitem{medcl_miccai23} Zhang Y, Li X, Chen H, et al. Continual learning for abdominal multi-organ and tumor segmentation[C]//International conference on medical image computing and computer-assisted intervention. Cham: Springer Nature Switzerland, 2023: 35-45.

\bibitem{medcl_miccai22_2} Zhang J, Xue P, Gu R, et al. Learning towards synchronous network memorizability and generalizability for continual segmentation across multiple sites[C]//International Conference on Medical Image Computing and Computer-Assisted Intervention. Cham: Springer Nature Switzerland, 2022: 380-390.

\bibitem{medcl_tmi23} Zhang J, Gu R, Xue P, et al. S 3 R: Shape and semantics-based selective regularization for explainable continual segmentation across multiple sites[J]. IEEE Transactions on Medical Imaging, 2023, 42(9): 2539-2551.

\bibitem{medcl_media_24} Zhu Z, Ma X, Wang W, et al. Boosting knowledge diversity, accuracy, and stability via tri-enhanced distillation for domain continual medical image segmentation[J]. Medical image analysis, 2024, 94: 103112.

\bibitem{clseg_survey} Yuan B, Zhao D. A survey on continual semantic segmentation: Theory, challenge, method and application[J]. IEEE Transactions on Pattern Analysis and Machine Intelligence, 2024, 46(12): 10891-10910.

\bibitem{medcl_survey} Kumari P, Chauhan J, Bozorgpour A, et al. Continual learning in medical image analysis: A comprehensive review of recent advancements and future prospects[J]. Medical Image Analysis, 2025: 103730.

\bibitem{general_survey} Wang L, Zhang X, Su H, et al. A comprehensive survey of continual learning: Theory, method and application[J]. IEEE transactions on pattern analysis and machine intelligence, 2024, 46(8): 5362-5383.



\end{thebibliography}
\end{document}